\documentclass[]{fairmeta}

\usepackage{amsfonts}
\usepackage{amsmath}
\usepackage{amssymb}
\usepackage{nicefrac}
\usepackage{algorithm}
\usepackage{algorithmic}
\usepackage{enumitem}
\usepackage{makecell}

\newcommand{\method}{\textsc{TreeFlex}}
\newcommand{\react}{\textsc{ReAct}}
\newcommand{\foldagent}{\textsc{FoldAgent}}
\newcommand{\tool}[1]{\texttt{#1}}
\lstnewenvironment{systemprompt}{%
  \lstset{%
    basicstyle=\ttfamily\fontsize{7}{7.3}\selectfont, breaklines=true, breakatwhitespace=true, columns=fullflexible, keepspaces=true, showstringspaces=false, frame=l, framesep=6pt, xleftmargin=1em, aboveskip=4pt, belowskip=4pt, literate={—}{{\textemdash}}1 }%
}{}

\titleformat*{\paragraph}{\normalfont\bfseries}

\title{TreeFlex:\\ Agents Navigating Their Own Search Trees}

\author{Dongzhuo Li}
\author{Fangzhou Cheng}
\author{Weiwei Li}
\author{Kevin Han}
\author{Haixu Ma}
\author{Zhongchang Sun}
\author{Ruiyi Li}
\author{Serena Li}
\author{Xiangjun Fan}
\author{Lizhu Zhang}
\author[\dagger]{Yuandong Tian}
\affiliation{Meta AI}
\contribution[\dagger]{Work done while at Meta}
\metadata[Correspondence]{Dongzhuo Li at \email{lidongzh@meta.com}}

\abstract{ A ReAct-style agent accumulates observations, intermediate results, and failed attempts in a single context. As this context grows, stale and off-path information consumes its limited capacity and obscures relevant evidence, making focused reasoning increasingly difficult. We introduce \method{}, a framework in which the agent's own policy decides where to search. The agent records each intermediate state, with a short summary, as a node in a search tree maintained by the harness. The policy is a single language model that acts in two roles. In the Reflection role, it has a global view of the tree through these summaries and picks a node to resume, together with an instruction. In the Execution role, the harness restores the path that led to that node, and the model carries out the instruction with the usual task tools until it saves a new node or proposes a solution, which Reflection then approves or sends back to another node. Through these choices, the policy decides whether to deepen the current line, branch from or revisit an earlier state, verify a proposal, or stop, without a fixed traversal order or a score for each node. We train navigation, task actions, checkpointing, and proposal review jointly with a tree-aware adaptation of GRPO, using task-level outcome rewards augmented by process rewards. After small-scale RL of a 4B model, with the same active-context limit for all methods, \method{} reaches 48.8\% accuracy on a challenging deep-research benchmark, versus 41.5\% for flat \react{} and 42.8\% for \foldagent{}, and its Pass@8 is about ten points higher than either. On SWE-bench Verified, with all three systems trained with RL from the same 9B model, \method{} resolves 54.9\% of tasks, versus 43.3\% for \foldagent{} and 45.2\% for its step-wise variant. }

\begin{document}

\maketitle

\section{Introduction}
\label{sec:introduction}

On long-horizon tasks, an agent rarely reaches a solution in one straight path. Several candidate solutions may build on the same partial progress, a failed attempt may still reach a state worth resuming from, and later evidence may make an abandoned direction worth revisiting. In deep research, an agent may check several candidate answers against the same evidence and return to an earlier one that it had dismissed when a later document changes the picture. In agentic software engineering (SWE), it may try several fixes for the same reproduced bug and return to an earlier repository state when a later test implicates code it had set aside. Yet a standard \react{} agent~\citep{yao2023react} keeps all of this in one linear context. Every later model call then carries abandoned attempts and long off-path observations along with the evidence it needs, and undoing an attempt may also require undoing its effects on the environment. One remedy is to compress the history. Learned summarization and memory-consolidation methods select what past information remains available~\citep{zhou2025mem1,li2026mempo}, but whatever they discard is gone for the rest of the episode. We instead keep several intermediate states that the agent can return to, and show the model only the context of the path it is currently working on.

Tree-search methods for language models make alternative states explicit~\citep{yao2023tree,hao2023reasoning, zhou2024lats}, but having a tree does not tell the agent which branch to work on next. A hand-designed DFS, BFS, or MCTS controller imposes its own selection rule. A learned score for each node is more adaptive, but it evaluates each node in isolation, while its usefulness changes as other branches reveal new evidence and the search approaches its limits. A score also leaves open how to continue from the selected state. In other words, the useful decision is often a node paired with a plan for what to do there, not a node alone. We therefore ask whether the model that does the task can also make this choice, based on what the search has found so far.

A second question is how to train this behavior end to end. Standard agentic RL treats each episode as one growing token sequence with one final reward. A tree search breaks this assumption, because the model makes decisions from different saved states and therefore sees a different history each time. Concatenating all branches into one sequence would show the model histories it never saw during the search and bring back the long-context problem.

\begin{figure}[t]
  \centering
  \includegraphics[width=\textwidth]{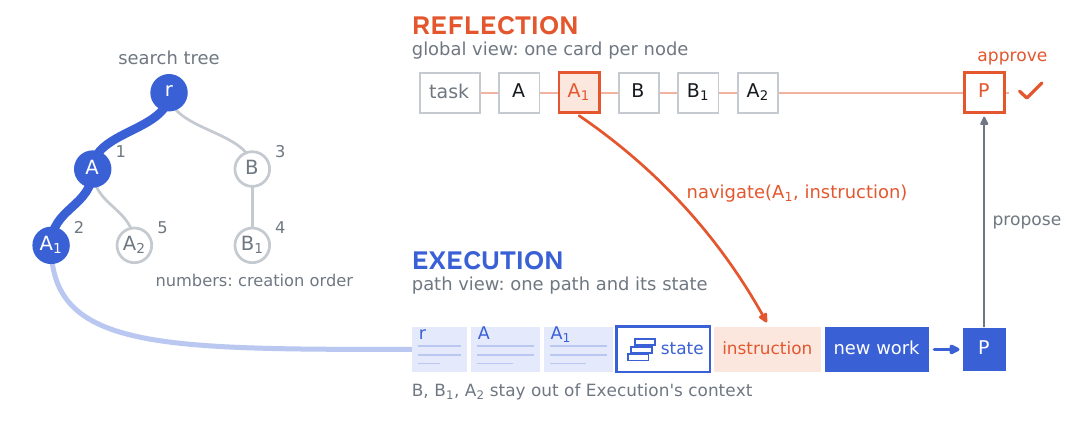}
  \caption{\textbf{One TreeFlex search.} Left: the search tree, numbered in creation order. Top: Reflection sees the task and a short card for every node. Although $B$, $B_1$, and $A_2$ are newer, it calls \tool{navigate} to resume $A_1$ with a new instruction. Bottom: the harness restores only the path $r\!\rightarrow\!A\!\rightarrow\!A_1$ and its saved state for Execution, so the other branches stay out of its context. Execution proposes $P$, and Reflection then approves it. Both roles use the same model. In deep research, the restored path consists of cards and the saved state is empty. In SWE, the restored path is the raw context along the path, and the saved state is a repository snapshot.}
  \label{fig:overview}
\end{figure}

We introduce \method{}, a framework that saves the agent's intermediate states as nodes of a search tree, each summarized by the model itself. The model works in two roles, Reflection and Execution (we later use each role's name for the model acting in that role). In Reflection, it receives the original task and a short summary of every node, which we call its card, and selects one node together with a focused instruction. The harness then restores the path to that node for Execution, where the model uses the usual task tools to carry out the instruction and either saves a new checkpoint or proposes a final solution. Reflection then either approves the proposal, which ends the search, or rejects it by selecting one node to continue from. Continuing from the current node deepens a line, while returning to an earlier node or the root starts a new branch from that state. In classical search terms, Reflection manages the frontier, which is the set of states from which search can continue. For example, best-first search keeps this frontier as an open list ordered by priority. However, unlike such a list, Reflection can return to states it has already expanded, and it pairs each choice with an instruction. Figure~\ref{fig:overview} shows one such search.

This design lets the search grow without any context having to hold every branch. Reflection's record grows by only one short card per node, and each Execution segment only sees one path. A node saves the environment state, such as a repository snapshot, as well as the raw prefix or a summary card for the segment, and returning to an earlier node restores the workspace. Therefore, the agent can use more compute to deepen a line, start a sibling, revisit earlier work, or verify a proposal without adding every explored branch to later Execution contexts.

Although the tree branches, the model's calls still happen one after another. We use this property to adapt GRPO~\citep{shao2024deepseekmath}. Each complete search counts as one rollout with one task reward, and each model output is trained on the exact context it was generated from, so no tree-shaped attention mask is needed.

We study \method{} on deep research with BrowseComp-Plus~\citep{chen2025browsecompplus} and on SWE with SWE-bench Verified~\citep{jimenez2024swebench,openai2024swebenchverified}. On the BrowseComp-Plus test split used by \foldagent{}~\citep{sun2025foldagent}, we compare it with flat \react{}, which keeps the whole history in one context, and \foldagent{} under the same model and resource limits. Before RL, \method{} has the highest accuracy at both 4B and 30B. After RL training of Qwen3-4B-Instruct-2507~\citep{qwen2025qwen3} with a 32k active-context limit, \method{} reaches 48.8\% accuracy, versus 41.5\% for flat \react{} and 42.8\% for \foldagent{}, and 66.0\% Pass@8, versus 56.0\% and 55.3\%. On SWE, we train \method{} and \foldagent{} from Qwen3.5-9B with RL on SWE-Gym tasks~\citep{pan2025swegym} and select each run's training step on separate SWE-Gym validation tasks. \method{} resolves 54.9\% of tasks, versus 43.3\% for \foldagent{} and 45.2\% for a \foldagent{} variant with a step-wise delegation prompt that we wrote.

Our contributions are:

\begin{itemize}[leftmargin=*,itemsep=2pt]
  \item \textbf{Policy-directed tree search.} One model decides when to continue, branch, backtrack, verify, or stop, while the harness only saves and restores contexts and environment states.
  \item \textbf{Tree-aware policy optimization.} We adapt GRPO to treat a complete tree search as one sampled rollout, train each model output on the exact context it was generated from, and combine the task reward with process penalties on the actions that caused them.
  \item \textbf{Evaluation across two agent domains.} On deep research, we compare three ways of organizing an agent's history, a single linear context (\react{}), context folding (\foldagent{}), and a search tree (\method{}), under the same model and context limit. On SWE, we compare the complete \foldagent{} and \method{} systems. We report task success and resource use.
\end{itemize}

\section{TreeFlex}
\label{sec:method}

\subsection{Agent Design}
\label{sec:setting}

\paragraph{Search tree.}

Let $x$ be a task and $\mathcal{D}$ its ordinary domain tools. \method{} maintains a rooted tree $\mathcal{T}$ whose root $r$ stores the task prompt and the initial environment. Every other node $v$ is saved at the end of an Execution segment, either by \tool{checkpoint} or, for a proposed final solution, by \tool{commit}. The segment starts from an existing node $p_v$, its parent, and carries out instruction $I_v$. A node thus corresponds to a whole segment of work, which usually spans many turns. It stores everything needed to resume from it:

\begin{equation}
v = (p_v,\ I_v,\ s_v,\ m_v,\ e_v),
\end{equation}

where $s_v$ is a model-authored card, $m_v$ the context state, and $e_v$ the environment state, such as a repository snapshot in SWE.

\paragraph{Roles and tree tools.}
\label{sec:operations}

The language model acts in two roles that differ only in what they see and which tools they can call (Table~\ref{tab:operations}). In Reflection, it sees the original task and a global record with one entry per node, which gives the node's identifier, parent, and card but not the raw context of any segment. In Execution, it sees the \emph{path view}: the context restored along the path from the root to the selected node, the selected node's environment state $e_v$, and the new instruction, but nothing from other branches. In deep research, retrieved evidence can be cited in a card and fetched again, so the restored context consists only of the cards along the path. In SWE, error traces and diffs are hard to summarize faithfully, so the restored context is the raw context along the path. When it no longer fits, the harness stores a compact resumable context containing the path cards and a truncated repository diff (Appendix~\ref{app:context-layouts}).

This division of work follows from what each role can see. Execution has the detailed evidence of one path, so it does the work and proposes solutions. Reflection sees every branch through short cards, so it chooses where to work next and checks proposed solutions against evidence from all branches. It cannot edit a proposal, because it lacks the raw evidence.

\begin{table}[h]
  \centering
  \caption{\textbf{Tree tools of Reflection and Execution.} Deep research calls \tool{commit} with an answer and an explanation. SWE exposes it as OpenHands' \tool{finish(message)}, which proposes the current repository state.}
  \label{tab:operations}
  \small
  \begin{tabular}{p{0.29\textwidth}p{0.12\textwidth}p{0.51\textwidth}}
    \toprule
    \textbf{Tool} & \textbf{Role} & \textbf{What it does} \\
    \midrule
    \tool{navigate(node, instruction)} & Reflection & Choose a node to resume and write an instruction for the next segment. Creates no node. \\
    Domain tools $\mathcal{D}$ & Execution & Work on the task, such as retrieving documents or editing and testing code. \\
    \tool{checkpoint(summary)} & Execution & Save the current state as a new child of the node where the segment started, with \texttt{summary} as its card, and return the card to Reflection. \\
    \tool{commit(solution)} & Execution & Save the current state as a new child of the node where the segment started, holding a proposed final solution for Reflection to review. \\
    \tool{approve\_commit(node)} & Reflection & Accept the proposed solution at \texttt{node} unchanged, which ends the search. \\
    \bottomrule
  \end{tabular}
\end{table}

\paragraph{Search procedure.}
\label{sec:context}

A search alternates between the two roles. Reflection's first \tool{navigate} call must choose the root. Each Execution segment ends with \tool{checkpoint} or \tool{commit}. If a segment runs out of context, the harness requires a checkpoint. The resulting card or proposal returns as the result of the pending \tool{navigate} call. Reflection thus works like a ReAct loop over segment results (Appendix~\ref{app:context-layouts}) and reconsiders where to search after every segment.

Reflection's choices shape the tree. Resuming the latest node deepens the current line, resuming an earlier node starts a sibling branch, and resuming the root starts a new line. The search can therefore behave like depth-first, breadth-first, or best-first search, and the model learns when to do which, without a fixed rule. Segments run one at a time, without parallel model calls.

The search ends when Reflection accepts a proposed solution or when the rollout reaches its turn limit. If Reflection navigates instead, the search continues from the chosen node, while the proposal remains available for later approval. The harness validates calls and manages saved state, but does not choose where to navigate. Appendices~\ref{app:context-limits} and~\ref{app:runtime-procedure} give the settings and pseudocode.

\subsection{RL Training}
\label{sec:training}

\paragraph{Rollouts.} In standard agentic RL, a rollout is one growing token sequence. In contrast, a \method{} search or rollout is split across several sequences, because Reflection keeps its own context and each Execution segment starts from the saved context of its selected node. Training on such a search raises two questions: what counts as one sample and which context each output is trained on. For the first, we treat each complete search as one rollout and one member of its GRPO group, since only the complete search receives a task reward. The task reward therefore applies to every output in the search, whereas a process penalty applies only to the turn it concerns. For the second, each output is trained on exactly the context it was generated from. Its log-probability is computed given that context, and the restored context is used only as input, so the loss applies only to tokens the model newly generated.

\begin{figure}[t]
  \centering
  \includegraphics[width=\textwidth]{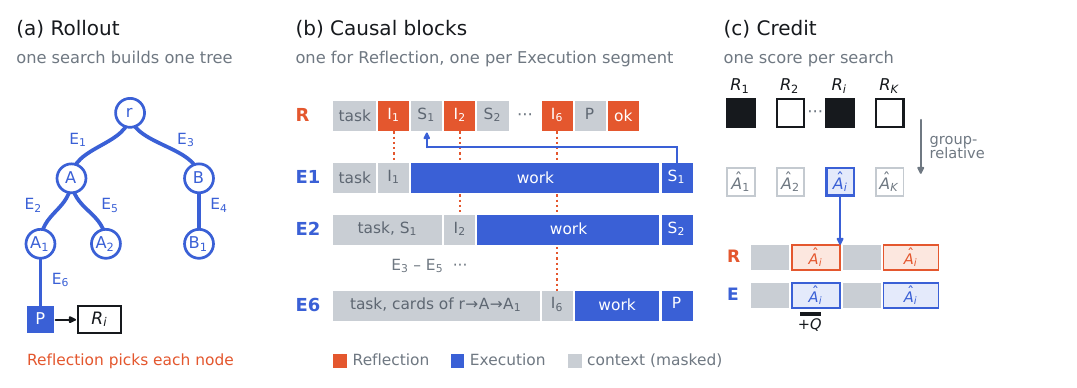}
  \caption{\textbf{Tree-aware GRPO on one search.} \textbf{(a)} The search from Figure~\ref{fig:overview}. In Reflection, the policy picks each node to resume. In Execution, it saves one checkpoint per segment $E_j$ and, in $E_6$, proposes $P$. The proposal $P$ receives the terminal score $R_i$. \textbf{(b)} Reflection forms one causal block ($R$), and each Execution segment forms its own ($E_j$). Colored spans are the model's own outputs, which are trained, and gray spans are masked context. Instruction $I_j$ is trained as a Reflection output in $R$ and appears only as context in $E_j$ (dotted). Card $S_j$ is trained as an Execution output in $E_j$ and appears only as context when it returns to $R$. The blocks follow the deep-research setting, in which Execution sees the cards along its path and only its own instruction. \textbf{(c)} Each of the $K$ searches for a task receives one score. Its group-relative advantage $\hat A_i$ applies to every generated token of search $i$, and a process penalty $Q$ applies only to the turn it concerns. The objective uses a token, block, or rollout mean and can mix in a role mean with weight $\lambda$.}
  \label{fig:training-objective}
\end{figure}

\paragraph{Causal blocks and loss mask.} A \emph{causal block} is a maximal sequence over which context grows only by appending tokens. It therefore preserves the exact prefix from which each generated token was sampled. Reflection's context grows throughout the search, so Reflection forms a single block. An Execution segment ends with \tool{checkpoint} or \tool{commit}, and the next segment starts from a restored context, so each segment forms its own block. A segment's tokens are therefore trained only in its own block, even when several branches build on it (Figure~\ref{fig:training-objective}). For rollout $i$, block $b$, and token position $t$, let $z_{i,b,t}$ be the token, $h_{i,b,t}$ the context that it was generated from, and $m_{i,b,t}\in\{0,1\}$ its loss mask. The mask is one for tokens that the model generated in block $b$, and zero for everything else, namely the task prompt, restored context, harness messages, and tool observations. Instructions or summary cards that are copied into the other role's context are also masked. In other words, an instruction is trained once, in the Reflection turn that wrote it, and not again when Execution reads it. Likewise, a card is trained in the Execution segment that wrote it, not when Reflection reads it.

\paragraph{Rewards and advantages.} For task $x$, the old policy samples $K$ complete searches. Search $i$ ends with a final output $y_i$, the submitted answer in deep research or the graded repository state in SWE, and receives a binary terminal score $R_i=R(x,y_i)$. We use only non-positive process rewards, which we call process penalties. In successful searches, some generated turns receive a process penalty $Q_{i,b,t}$, which applies only to the tokens of that turn. The training signal has three components, and the ablations study a fourth.

\begin{itemize}[leftmargin=*,itemsep=2pt,topsep=3pt]
  \item \textbf{Terminal task reward.} In deep research, a search scores one if the precision-first judge accepts its submitted answer. In SWE, it scores one if the official tests mark the task as resolved. Every other outcome scores zero.
  \item \textbf{Failure penalty.} Following \foldagent{}~\citep{sun2025foldagent}, a turn that issues a malformed or rejected tool call receives $Q=-1$.
  \item \textbf{Out-of-scope (OOS) penalty.} In deep research, we also penalize work outside the assigned instruction. A judge sees the original task, the selected node's card, the instruction, and brief metadata for each retrieval action. If the judge labels an action out of scope, the turn containing it receives $Q=-0.2$. Checkpoint and proposal turns are never penalized, and uncertain, malformed, or missing judge outputs give no penalty.
  \item \textbf{Reflection finalization penalty (ablations only).} In our design, it is Execution that decides when to propose a solution, so Reflection should not use its instruction to request one. An instruction that directs Execution to produce the solution rather than new evidence, such as ``Propose Candidate A as the answer'', receives $Q=-1$. A pattern rule or the OOS judge detects such instructions (Appendix~\ref{app:reward}).
\end{itemize}

There is no separate reward for checkpointing, selecting a particular node, or favoring any tree shape. Search structure is therefore not prescribed by any tree-specific reward.

We use the terminal rewards and process penalties to compute group-relative advantages. The group mean $\bar R_x$ and standard deviation $s_x$ are computed from the $K$ terminal scores alone, so a penalty on one turn does not move the baseline of any other turn. Following \foldagent{}~\citep{sun2025foldagent}, we divide by the standard deviation:

\begin{equation}
\widehat A_{i,b,t}
= \frac{\operatorname{clip}(R_i+Q_{i,b,t},0,1)-\bar R_x}{s_x}.
\label{eq:treeflex-advantage}
\end{equation}

When all $K$ searches have the same score, we divide by one instead, so a group in which every search succeeds keeps its penalties at their original scale. Without penalties, every generated token of a search receives the same advantage.

\paragraph{Loss aggregation.} Let $\rho_u(\theta)$ be the importance ratio of a generated token $u=(i,b,t)$ under the context $h_{i,b,t}$ it was generated from. The objective is a weighted sum of PPO's clipped surrogate~\citep{schulman2017ppo} over the generated tokens $\mathcal{U}$ of one PPO minibatch:
\begin{equation}
J_{\method{}}(\theta)=\mathbb{E}\Big[\sum_{u\in\mathcal{U}} w_u\,
  \min\big(\rho_u(\theta)\widehat A_u,\
  \operatorname{clip}(\rho_u(\theta),1-\epsilon_{\mathrm{low}},1+\epsilon_{\mathrm{high}})\,\widehat A_u\big)\Big],
\label{eq:treeflex-objective}
\end{equation}
where $w_u\ge 0$ and $\sum_{u\in\mathcal{U}}w_u=1$, and the clipping range is decoupled as in DAPO~\citep{yu2025dapo}. For negative advantages, the per-token term is also bounded below by $c\,\widehat A_u$ (dual clipping with $c=3$; Appendix~\ref{app:hyperparams}). A rollout spans several blocks, and both blocks and rollouts vary widely in length. The choice of $w_u$ decides how much each part of a rollout counts. Write $k(u)$, $i(u)$, and $r(u)$ for the block, rollout, and role of token $u$, and $\mathcal{U}_k$, $\mathcal{U}_i$, and $\mathcal{U}_r$ for the generated tokens of block $k$, rollout $i$, and role $r$. We consider three averages, which differ in the unit that receives equal weight.

\begin{itemize}
\item \textbf{Token mean.} Set $w_u=1/|\mathcal{U}|$, which averages all tokens together. Every token counts the same, so a rollout that generates ten times as many tokens counts ten times as much.
\item \textbf{Block mean.} Set $w_u=1/(|\mathcal{K}|\,|\mathcal{U}_{k(u)}|)$, with $\mathcal{K}$ the blocks in the minibatch. It first averages the tokens within each block and then averages over blocks. Every block counts the same. This is the per-sequence average of common RL trainers, applied to our blocks. A rollout's weight grows with its number of blocks, and a short Execution segment counts as much as a long one. A block too long for one training row is split, and each part then counts as a block (Appendix~\ref{app:training-details}).
\item \textbf{Rollout mean.} Set $w_u=1/(|\mathcal{I}|\,|\mathcal{U}_{i(u)}|)$, with $\mathcal{I}$ the rollouts in the minibatch. It first averages the tokens within each rollout and then averages over rollouts. This extends the per-sample averaging of the original GRPO objective~\citep{shao2024deepseekmath} to rollouts containing multiple blocks, giving each rollout equal total weight regardless of its length or number of blocks.
\end{itemize}

A separate imbalance arises between roles. Execution produces many more tokens than Reflection, so Reflection's decisions about where to search and whether to accept a proposal receive little weight. The role mean, $w^{\mathrm{role}}_u=1/(|\mathcal{R}_+|\,|\mathcal{U}_{r(u)}|)$ with $\mathcal{R}_+$ the roles present, first averages within each role and then across roles, giving each role equal weight. A weight $\lambda\in[0,1]$ mixes the role mean with one of the three averages above, $w_u=(1-\lambda)\,w^{\mathrm{base}}_u+\lambda\,w^{\mathrm{role}}_u$. We train \method{} with the rollout mean and $\lambda=0$ in both domains. Section~\ref{sec:ablations} compares it with the other averages and with $\lambda=0.3$. Appendices~\ref{app:training-details} and~\ref{app:bcp-judges} give the remaining details.


\section{Experimental Setup}
\label{sec:setup}
\label{sec:experiments}

\subsection{Tasks and Datasets}

\paragraph{Deep Research.}
\label{sec:browsecomp}

We evaluate deep research with BrowseComp-Plus, whose compositional questions require finding and synthesizing evidence across multiple documents in a frozen corpus~\citep{chen2025browsecompplus}. We use the train/test split of \citet{sun2025foldagent}, with 680 training and 150 test questions, and further hold out 64 of the training questions as a fixed validation split, leaving 616 for training.

\paragraph{Agentic SWE.}
\label{sec:swebench}

We draw 600 executable Python repository tasks from SWE-Gym~\citep{pan2025swegym}, spanning ten repositories, and split them into 500 training tasks and 100 validation tasks. Each task pairs a natural-language issue with a repository snapshot and hidden executable tests. We train each method on tasks on which earlier training rollouts had mixed outcomes. \foldagent{} trains on the 217 tasks on which its own rollouts had mixed outcomes, and \method{} on the 250 tasks on which the rollouts of either method had mixed outcomes. Evaluation uses all 500 SWE-bench Verified tasks and the official grader~\citep{jimenez2024swebench,openai2024swebenchverified}. Appendix~\ref{app:data-splits} describes how the training tasks were selected.

\subsection{Comparison Protocol}
\label{sec:exp-setup}

Across both domains, comparisons between \foldagent{}~\citep{sun2025foldagent} and \method{} use a 32,768-token active-context limit. \foldagent{} may launch at most ten branches, and \method{} may make at most ten navigations. Within these limits, each method keeps its own context layout, prompts, state management, and tool interfaces.

\paragraph{BrowseComp-Plus comparison.} On BrowseComp-Plus, we compare flat \react{}, \foldagent{}, and \method{} with the same base model, retriever, and retrieval limits. We evaluate the released Qwen3-4B-Instruct-2507 and Qwen3-30B-A3B-Instruct-2507 models~\citep{qwen2025qwen3} without task-specific RL, and train all three methods from the 4B model.

Flat \react{} uses the same 32,768-token limit, except in one diagnostic run with a 254k-token limit. Appendix~\ref{app:context-limits} gives the exact accounting.

Following \citet{sun2025foldagent}, we train all three methods on the same data for 50 steps, using 32 questions per step, eight rollouts per question, and a learning rate of $10^{-6}$. The methods differ in their training objectives: \method{} uses tree-aware GRPO, \foldagent{} uses FoldGRPO, and flat \react{} uses GRPO. We make four changes to the original setup. We use a 4B rather than a 36B model, select each run's training checkpoint on the validation split, and sample eight rollouts per test question rather than one greedy rollout. We also apply the 32k active-context limit to the context after the initial prompt and use a larger model window so that the next call still fits. The answer judge of Appendix~\ref{app:bcp-judges} scores every rollout, and Appendix~\ref{app:hyperparams} gives the full configuration.

\paragraph{SWE-bench Verified comparison.} On SWE-bench Verified, we compare the complete \foldagent{} and \method{} systems. Both start from the same Qwen3.5-9B model with thinking enabled~\citep{qwen2026qwen35}. On their respective task subsets drawn from the pool described above, we train \method{} with tree-aware GRPO and \foldagent{} with FoldGRPO. Both use 32 tasks per step, eight rollouts per task, and a learning rate of $10^{-6}$. Each run trains for 25 steps, about three to four epochs over its tasks, and we select the checkpoint with the highest validation Mean@3 among steps 5, 10, 15, 20, and 25. For a controlled comparison, we also match the repository images, evaluation turn and time limits, sampling parameters, and official grader. On SWE, \method{} also allows at most two children per node. We sample three rollouts per test task and grade each on one repository state, whether or not the agent called \tool{finish}. For \foldagent{}, this is its final workspace. For \method{}, it is the approved proposal if there is one, otherwise the latest proposal, and otherwise the final workspace. Appendix~\ref{app:swe-alignment} gives the full alignment, including the remaining differences in training rewards, and Appendix~\ref{app:reward} specifies all rewards.

\section{Results}
\label{sec:results}

All comparisons are paired over the same questions or tasks. We call a difference significant when its 95\% paired-bootstrap confidence interval excludes zero (Appendix~\ref{app:paired}).

\subsection{Deep Research}

\paragraph{Main results.}

\method{} gives the highest mean accuracy for every model under the same context limit (Table~\ref{tab:main-results}).

\begin{table}[t]
  \centering
  \caption{\textbf{TreeFlex gives the highest mean accuracy under matched context limits.} Each cell aggregates eight samples for each of 150 questions. Mean@8 is the accuracy averaged over the eight samples, and Pass@8 counts a question as solved if any sample is correct. Usage values are means per rollout. Generated tokens exclude retrieved observations, and reads count \tool{get\_document} and \tool{open\_page} calls. Ctx. is the active-context limit after the initial prompt, and Br./nav. is the maximum number of branches (\foldagent{}) or navigations (\method{}). The flat \react{} row with a 254k limit is a diagnostic outside the matched comparison.}
  \label{tab:main-results}
  \small
  \setlength{\tabcolsep}{4.5pt}
  \begin{tabular}{llccrrrrr}
    \toprule
    & & & & \multicolumn{2}{c}{\textbf{Quality (\%)}} & \multicolumn{3}{c}{\textbf{Usage per rollout}} \\
    \cmidrule(lr){5-6}\cmidrule(lr){7-9}
    \textbf{Model} & \textbf{Method} & \textbf{Ctx.} & \textbf{Br./nav.} & \textbf{Mean@8} & \textbf{Pass@8} & \textbf{Gen. tokens} & \textbf{Searches} & \textbf{Reads} \\
    \midrule
    \multirow{3}{*}{Qwen3-30B (base)}
      & Flat \react{} & 32k & 0 & 42.6 & 55.3 & 584 & 3.1 & 3.7 \\
      & \foldagent{} & 32k & 10 & 46.1 & \textbf{66.7} & 3,212 & 8.7 & 11.5 \\
      & \method{} & 32k & 10 & \textbf{48.3} & 62.7 & 4,833 & 14.3 & 20.6 \\
    \midrule
    \multirow{3}{*}{Qwen3-4B (base)}
      & Flat \react{} & 32k & 0 & 36.2 & 56.0 & 1,082 & 2.4 & 2.3 \\
      & \foldagent{} & 32k & 10 & 33.1 & 54.0 & 2,553 & 5.7 & 12.3 \\
      & \method{} & 32k & 10 & \textbf{43.6} & \textbf{65.3} & 4,493 & 8.8 & 15.6 \\
    \midrule
    \multirow{4}{*}{Qwen3-4B (RL)}
      & Flat \react{} & 32k & 0 & 41.5 & 56.0 & 829 & 3.4 & 2.0 \\
      & \foldagent{} & 32k & 10 & 42.8 & 55.3 & 1,990 & 9.3 & 3.5 \\
      & \method{} & 32k & 10 & \textbf{48.8} & \textbf{66.0} & 5,677 & 20.7 & 9.7 \\
      \cmidrule(lr){2-9}
      & Flat \react{} & 254k & 0 & 42.7 & 54.7 & 1,501 & 5.5 & 2.5 \\
    \bottomrule
  \end{tabular}
\end{table}

After RL, \method{} reaches 48.8\% Mean@8 and 66.0\% Pass@8, significantly outperforming flat \react{} by 7.3 and 10.0 points and \foldagent{} by 6.0 and 10.7 points, respectively. An RL-trained flat \react{} diagnostic with a 254k-token limit reaches only 42.7\% Mean@8 and 54.7\% Pass@8. The run does use the extra room, with 12\% of its late-training rollouts exceeding 64k context tokens and a 90th percentile of 87.1k, versus 33.3k for the matched 32k run, yet its Mean@8 improves by only 1.2 points over the matched 32k run (Appendix~\ref{app:context-limits}).

Without task-specific RL, \method{} gives the highest Mean@8 at both 4B (43.6\%) and 30B (48.3\%), although \foldagent{} has the highest 30B Pass@8 (66.7\%). On validation, \method{} leads all baseline runs at every evaluated training step (Appendix~\ref{app:training-dynamics}).

\begin{figure}[t]
  \centering
  \includegraphics[width=\textwidth]{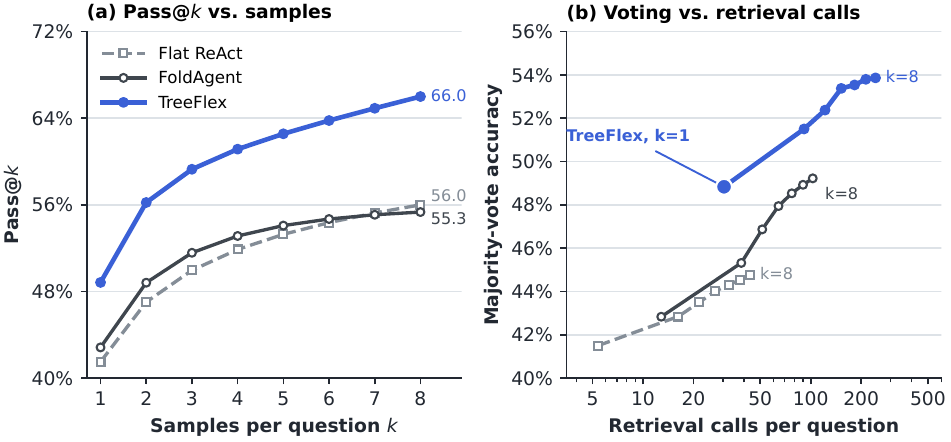}
  \caption{\textbf{Eight samples of a baseline do not close the Pass@$k$ gap.} Curves use the 32k Qwen3-4B RL models and eight samples for each of 150 questions. \textbf{(a)} Pass@$k$. \textbf{(b)} Majority-vote accuracy over normalized answers against retrieval calls (searches plus reads), averaged over all $k$-sample subsets with ties broken at random. Voting never uses correctness labels. We omit $k=2$ because it gives the same result as $k=1$.}
  \label{fig:majority-voting}
\end{figure}

\paragraph{Repeated sampling and voting.} These improvements require more inference. \method{} uses 5.7k generated tokens and 30.4 retrieval calls per rollout, compared with 2.0k and 12.8 for \foldagent{}, and 0.8k and 5.4 for flat \react{}.

Figure~\ref{fig:majority-voting} asks whether more samples let a baseline catch up. With the same number of samples, they do not. At $k=8$, \method{} reaches 66.0\% Pass@8, versus 56.0\% for flat \react{} and 55.3\% for \foldagent{}. The paired outcomes are strongly asymmetric: \method{} solves 20 questions missed by all eight flat \react{} samples, while flat \react{} solves only 5 missed by all eight \method{} samples. The corresponding counts against \foldagent{} are 20 and 4. Repeated flat sampling mostly revisits the same evidence. Across its eight samples, flat \react{} opens only 5.3 distinct documents per question and reads 34\% of the annotated evidence documents, as much as a single \method{} rollout (34\%) (Appendix~\ref{app:evidence}). Majority voting likewise improves only slightly with repeated flat sampling. From one to eight samples, flat \react{} improves by only 3.3 points, reaching 44.8\%, whereas \method{} improves to 53.9\%, 9.1 points above eight-sample flat voting.

With a similar number of retrieval calls, a single \method{} rollout is more accurate than repeated sampling of flat \react{}. At 30.4 calls it reaches 48.8\%, 4.8 points above five-sample flat voting (44.0\% at 27.0 calls).

\paragraph{Accuracy by question difficulty.}

We next ask whether branching helps most on questions that a flat agent rarely solves, rather than uniformly improving easy questions. Following \citet{sun2025foldagent}, an independent flat \react{} run (Appendix~\ref{app:context-limits}) defines 58 easy questions (at least seven of eight samples correct), 67 hard questions (none correct), and 25 medium questions.

\begin{figure}[t]
  \centering
  \includegraphics[width=0.68\textwidth]{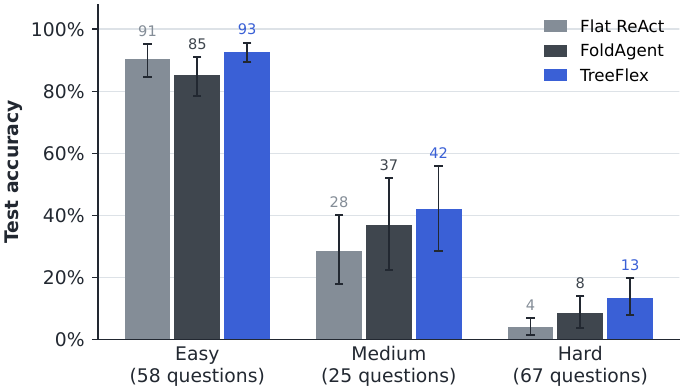}
  \caption{\textbf{TreeFlex leads in every difficulty bin.} Bars show mean accuracy over eight samples per question; error bars are 95\% question-bootstrap intervals. Difficulty follows the independent-reference rule of \citet{sun2025foldagent}.}
  \label{fig:difficulty}
\end{figure}

The improvements over flat \react{} are concentrated on harder questions (Figure~\ref{fig:difficulty}). On easy questions, \method{} reaches 92.7\%, close to flat \react{} (90.5\%) and above \foldagent{} (85.1\%). On medium questions, \method{} reaches 42.0\%, versus 28.5\% for flat \react{} and 37.0\% for \foldagent{}. On hard questions, it reaches 13.4\%, versus 3.9\% and 8.4\%. \method{} also searches more on harder questions. From easy to hard, searches rise from 8.9 to 29.9, reads from 7.6 to 11.8, and tree size from 5.8 to 9.1 nodes, whereas flat \react{} stays below four searches and about two reads in every bin.

\subsection{Agentic SWE}

Table~\ref{tab:swebench} compares the two systems before and after RL. It also includes a \foldagent{} variant with a step-wise delegation prompt, described below. Without task-specific RL, \method{} reaches 51.1\% Mean@3 and 64.4\% Pass@3, versus 31.7\% and 44.0\% for \foldagent{}. RL raises \foldagent{} by 11.7 points, mostly by teaching it to finish within the context limit. The share of its rollouts that end with \tool{finish} rises from 4\% to 17\%, and rollouts that finish and resolve the task account for 9.7 of the 11.7 points. RL adds a smaller 3.8 points for \method{}, whose base policy already ends 86\% of rollouts with an approved proposal, so training has less to fix in how a search ends. After RL, \method{} resolves 54.9\% of tasks, significantly more than \foldagent{} (11.6 points) and its step-wise variant (9.7 points).

\begin{table}[t]
  \centering
  \caption{\textbf{TreeFlex resolves the most SWE-bench Verified tasks, before and after RL.} Each row uses Qwen3.5-9B with a 32k active-context limit and at most ten branches or navigations, and aggregates three rollouts for each of 500 tasks. Mean@3 is the resolve rate averaged over the three rollouts, and Pass@3 counts a task as resolved if any rollout resolves it. RL models are selected among steps 5, 10, 15, 20, and 25 by Mean@3 on 100 SWE-Gym validation tasks. Turns and minutes are means per rollout.}
  \label{tab:swebench}
  \small
  \setlength{\tabcolsep}{4.5pt}
  \begin{tabular}{llrrrr}
    \toprule
    & & \multicolumn{2}{c}{\textbf{Resolved (\%)}} & \multicolumn{2}{c}{\textbf{Usage per rollout}} \\
    \cmidrule(lr){3-4}\cmidrule(lr){5-6}
    \textbf{System} & \textbf{RL} & \textbf{Mean@3} & \textbf{Pass@3} & \textbf{Turns} & \textbf{Minutes} \\
    \midrule
    \multirow{2}{*}{\foldagent{}}
      & none & 31.7 & 44.0 & 34 & 5.0 \\
      & FoldGRPO & 43.3 & 57.8 & 49 & 6.4 \\
    \addlinespace
    \multirow{2}{*}{\shortstack[l]{\foldagent{}, step-wise\\delegation prompt}}
      & none & 49.5 & 63.0 & 132 & 21.9 \\
      & FoldGRPO & 45.2 & 58.2 & 96 & 10.0 \\
    \addlinespace
    \multirow{2}{*}{\method{}}
      & none & 51.1 & 64.4 & 98 & 24.8 \\
      & tree-aware GRPO & \textbf{54.9} & \textbf{68.0} & 113 & 29.3 \\
    \bottomrule
  \end{tabular}
\end{table}

With the released \foldagent{} prompt, the base Qwen3.5-9B model almost never launches a branch. Only 2 of its 1,500 rollouts on SWE-bench Verified do, so nearly every rollout keeps all its work in the main context, and 96\% of rollouts end at the context limit. We therefore wrote a step-wise delegation prompt that asks the main agent to split the repair into a sequence of small branches, each returning one checkable result such as one edit or one focused test run, and to choose each next branch from the report the previous one returns (Appendix~\ref{app:fa-microbranch-prompt}). This raises \foldagent{} to 49.5\% Mean@3, close to \method{}'s 51.1\%, although only 42\% of its rollouts end with \tool{finish} (Appendix~\ref{app:swe-finish}). RL on top of this prompt lowers its resolve rate to 45.2\%. During training, this variant receives a positive task reward only if the main agent calls \tool{finish}, whereas test grading evaluates the final workspace whether or not \tool{finish} is called. RL increases the share of rollouts that both finish and resolve the task from 31.5\% to 40.8\%, but reduces the share that resolve without \tool{finish} from 18.1\% to 4.4\%.

\method{} uses more turns and time per rollout. After RL, it takes 113 turns and 29.3 minutes, versus 49 and 6.4 for \foldagent{}. Before RL, the step-wise variant takes more turns than \method{} (132 versus 98) and a similar time (21.9 versus 24.8 minutes), and it is still less accurate.

\subsection{Training Objective Ablations}
\label{sec:ablations}

\paragraph{Deep research.} On BrowseComp-Plus, we vary three settings of the training objective: the loss aggregation of Equation~\eqref{eq:treeflex-objective} (token, block, or rollout mean), the role weight $\lambda$, which gives Reflection's few tokens more weight, and a finalization penalty, which discourages Reflection instructions that merely ask Execution to finish. Table~\ref{tab:lambda-ablation} lists seven variants. All other settings match the main configuration (Appendix~\ref{app:hyperparams}), including step selection on validation.

None of these settings is essential. All seven variants outperform flat \react{} (41.5\%) and \foldagent{} (42.8\%). The most accurate variant is the main configuration, which uses the rollout mean with $\lambda=0$ (48.8\% Mean@8). Each variant is a single training run, so we read the differences below as observations about these runs rather than as effects of the settings.

\begin{table}[t]
  \centering
  \caption{\textbf{Branching varies widely across variants while test accuracy stays within a few points.} Loss is the aggregation of Equation~\eqref{eq:treeflex-objective}, $\lambda$ the role weight, and Pen.\ the finalization penalty. Val. is Mean@3 on the 64 validation questions at the step selected for each run, and Mean@8 and Pass@8 are test results at that step. Tree behavior compares the first and last ten training steps, using 160 inspected rollouts from each. Branched is the share of rollouts in which some node has more than one child. Root and Ancestor nav. are the shares of navigations that target the root or an earlier ancestor. Internal fork is the share of rollouts in which a node other than the root has more than one child.}
  \label{tab:lambda-ablation}
  \small
  \setlength{\tabcolsep}{3.5pt}
  \begin{tabular}{lccrrrcccc}
    \toprule
    & & & \multicolumn{3}{c}{\textbf{Accuracy (\%)}} & \multicolumn{4}{c}{\textbf{Tree behavior during training (\%)}} \\
    \cmidrule(lr){4-6}\cmidrule(lr){7-10}
    \textbf{Loss} & $\lambda$ & \textbf{Pen.} & \textbf{Val.} & \textbf{Mean@8} & \textbf{Pass@8}
      & \textbf{Branched} & \textbf{Root nav.} & \textbf{Ancestor nav.} & \textbf{Internal fork} \\
    \midrule
    rollout (main) & 0 & no & 38.0 & \textbf{48.8} & \textbf{66.0}
      & 53 $\rightarrow$ 59 & 19 $\rightarrow$ 23 & 5 $\rightarrow$ 2 & 19 $\rightarrow$ 18 \\
    block & 0 & no & 34.9 & 48.4 & 65.3
      & 38 $\rightarrow$ 66 & 17 $\rightarrow$ 19 & 3 $\rightarrow$ 10 & 12 $\rightarrow$ 32 \\
    rollout & 0.3 & no & 35.4 & 45.3 & \textbf{66.0}
      & 40 $\rightarrow$ 51 & 13 $\rightarrow$ 9 & 5 $\rightarrow$ 5 & 16 $\rightarrow$ 26 \\
    token & 0.3 & no & \textbf{40.1} & 47.7 & \textbf{66.0}
      & 50 $\rightarrow$ 63 & 20 $\rightarrow$ 19 & 4 $\rightarrow$ 0 & 19 $\rightarrow$ 12 \\
    token & 0 & no & 33.9 & 46.7 & 65.3
      & 49 $\rightarrow$ 83 & 17 $\rightarrow$ 33 & 5 $\rightarrow$ 1 & 18 $\rightarrow$ 24 \\
    token & 0 & yes & 35.4 & 46.4 & \textbf{66.0}
      & 50 $\rightarrow$ 44 & 12 $\rightarrow$ 6 & 8 $\rightarrow$ 11 & 25 $\rightarrow$ 31 \\
    token & 0.3 & yes & 36.5 & 44.2 & 62.0
      & 47 $\rightarrow$ 71 & 19 $\rightarrow$ 22 & 3 $\rightarrow$ 4 & 14 $\rightarrow$ 30 \\
    \bottomrule
  \end{tabular}
\end{table}

The loss aggregation mainly changes the shape of the search. With $\lambda=0$, the rollout mean leaves the search much as it was at the start of training. The token mean makes the policy branch far more often and return to the root more often, and the block mean makes it fork more often below the root (Table~\ref{tab:lambda-ablation}). Yet the three variants end within 2.1 points of one another, so how often and where the policy branches does not predict its accuracy.

The role weight and the finalization penalty show no consistent pattern. Raising $\lambda$ to 0.3 lowers accuracy with the rollout mean but raises it slightly with the token mean, and its effect on branching reverses once the penalty is added. The penalty lowers the accuracy of the token mean at both values of $\lambda$.

\paragraph{Agentic SWE.} On SWE, RL improves the resolve rate over the base model only with the rollout mean (Table~\ref{tab:swe-loss}). With the same data, $\lambda=0$, and checkpoint-selection procedure, the token mean reaches 52.1\%, close to the base model's 51.1\%, whereas the rollout mean reaches 54.9\%, significantly above both.

\begin{table}[t]
  \centering
  \caption{\textbf{On SWE, the rollout mean improves over the base model and the token mean does not.} Both RL runs train on the 250-task union and use the step among 5, 10, 15, 20, and 25 with the highest validation Mean@3 (Val.). Test values aggregate three rollouts for each of 500 SWE-bench Verified tasks; $\Delta$ is the change from the base model. Returns is the share of test rollouts that return to the root or an earlier node.}
  \label{tab:swe-loss}
  \small
  \begin{tabular}{lrrrrrr}
    \toprule
    \textbf{Run} & \textbf{Step} & \textbf{Val.} & \textbf{Mean@3} & \textbf{Pass@3} & $\Delta$
      & \textbf{Returns} \\
    \midrule
    base model & -- & 22.3 & 51.1 & 64.4 & -- & 43.0 \\
    token mean & 20 & 26.7 & 52.1 & 67.4 & $+1.0$ & 24.7 \\
    rollout mean & 25 & \textbf{28.7} & \textbf{54.9} & \textbf{68.0} & $+3.8$ & 12.2 \\
    \bottomrule
  \end{tabular}
\end{table}

\FloatBarrier
\section{Related Work}
\label{sec:related}

\paragraph{Reasoning and acting.} \react{} interleaves language-model reasoning with environment actions in one context \citep{yao2023react}. Reflexion augments agents with persistent verbal feedback across attempts \citep{shinn2024reflexion}. These agents keep a single linear history. \method{} keeps this loop within each Execution segment but saves intermediate states in a tree that the policy can return to.

\paragraph{Tree search with learned models.} In game playing, AlphaZero and ELF OpenGo pair Monte Carlo tree search with a learned policy-value network~\citep{silver2018alphazero,tian2019elfopengo}. The search algorithm is fixed, and the network supplies priors and value estimates. Tree of Thoughts, Graph of Thoughts, world-model planning, and Language Agent Tree Search bring tree and graph search to language models~\citep{yao2023tree,besta2024graph,hao2023reasoning,zhou2024lats}. The language model proposes partial solutions and usually scores them, and a fixed procedure decides which to expand next. This procedure is breadth- or depth-first search in Tree of Thoughts, MCTS in world-model planning and Language Agent Tree Search, and a user-specified graph of operations in Graph of Thoughts. For web agents, \citet{koh2025tree} run a best-first search guided by a prompted value function and restore pages by resetting the environment and replaying actions. Policy-Guided Tree Search instead trains a separate graph-network policy with RL to decide whether to expand, branch, backtrack, or stop over the reasoning steps of an unmodified language model~\citep{li2025pgts}. Stream of Search, Searchformer, and Dualformer train models on serialized search traces~\citep{gandhi2024sos,lehnert2024searchformer,su2025dualformer}, and ASTRO trains on CoT traces converted from Monte Carlo tree search~\citep{kim2025astro}. In all four, the search is written out as one sequence, so returning to an earlier node means writing it out again rather than restoring a saved state. \method{} instead lets the agent's own policy choose which saved node to resume and what to do there, while the harness restores the corresponding context and environment state.

\paragraph{Learning information-seeking behavior.} RL has trained language models to interleave reasoning with search, as in Search-R1 and R1-Searcher~\citep{jin2025searchr1,song2025r1searcher}, and DeepResearcher scales this training to real web environments~\citep{zheng2025deepresearcher}. WebDancer and WebSailor fine-tune ReAct-style agents on synthesized trajectories and then train them with RL~\citep{wu2025webdancer,li2025websailor}, and WebSailor makes its training questions harder by obscuring key details. These systems learn strong search policies over one linear trajectory.

\paragraph{Policy optimization over trees.} Tree-GRPO uses tree-structured rollouts for finer credit assignment, but the trained agent still acts along one trajectory~\citep{ji2025treegrpo}. In \method{}, the tree is the agent's own search, and each complete search is one GRPO sample~\citep{shao2024deepseekmath}.

\paragraph{Learned context management.} At the reasoning-trace level, Accordion-Thinking learns when to summarize completed steps and discard their detailed token history~\citep{yang2026accordion}. MEM1 rewrites a compact internal state at every turn and discards the rest of its context, so its memory stays nearly constant in size~\citep{zhou2025mem1}. MemPO trains the agent to write its own memory summaries and rewards each summary by how much it raises the model's probability of the correct answer~\citep{li2026mempo}. For long-horizon web search, AgentFold learns at each step either to condense the latest step into a short summary or to merge it with recent summaries into a coarser one~\citep{ye2025agentfold}, and LongSeeker lets the agent summarize past steps, cut tool outputs down to exact excerpts, delete steps, or roll back to an earlier step~\citep{lu2026longseeker}. Context-Folding trains \foldagent{} to run subtasks in temporary branches and fold each back into the main context as a report~\citep{sun2025foldagent}. These approaches make compression or retention part of the agent's behavior, but keep only the current working history, so discarded work cannot be resumed later.

\paragraph{SWE agents.} SWE-bench evaluates whether an agent can resolve real repository issues under executable tests \citep{jimenez2024swebench}, while SWE-Gym supplies repository tasks and environments for training agents and verifiers~\citep{pan2025swegym}. OpenHands provides an open platform in which software agents act through CodeAct-style code execution~\citep{wang2024openhands}. SWE-agent designs the commands and feedback format for an agent that edits one workspace along a single ReAct-style trajectory~\citep{yang2024sweagent}. Agentless instead samples many candidate patches independently from a fixed pipeline of localization, repair, and patch validation, and filters them with tests~\citep{xia2025agentless}. SWE-Search runs Monte Carlo tree search over repository states kept in a git-like commit tree, selecting nodes with a hand-designed rule and scoring them with a prompted language model~\citep{antoniades2025swesearch}. In \method{}, the trained policy itself chooses which saved repository state to work from next.

\section{Limitations and Future Work}
\label{sec:discussion}

Our experiments are small in scale. We train a 4B model on 616 deep-research questions and a 9B model on at most 250 SWE tasks. Each training configuration is run once, so our significance tests capture question sampling but not variation across training runs. The comparisons are between complete systems that also differ in prompts and tool interfaces. \method{} uses more inference-time computation than the baselines, and its segments run sequentially, which adds latency.

Training at a larger scale would show whether these results hold for stronger models, more training data, and longer training. On SWE, validation accuracy was still rising at the last saved step, so longer training may help there directly. SFT on successful searches that save, compare, and revisit nodes could give RL a better starting point in both domains. Future work could also reward efficiency directly by penalizing generated tokens or tool calls in successful searches.

\section{Conclusion}
\label{sec:conclusion}

\method{} saves an agent's progress as nodes of a search tree, lets one model choose which node to resume and what work to do there, and trains both roles end to end with tree-aware GRPO. Across both domains, \method{} solves more tasks than baselines trained from the same starting model.

\bibliographystyle{assets/plainnat}
\bibliography{paper}

\beginappendix
\section{Method Details}
\label{app:implementation}

\subsection{Search Procedure}
\label{app:runtime-procedure}

Algorithm~\ref{alg:treeflex} gives the search loop of Section~\ref{sec:setting}.

\begin{algorithm}[t]
  \caption{One \method{} search}
  \label{alg:treeflex}
  \begin{algorithmic}[1]
    \STATE Initialize tree $\mathcal{T}$ with root $r$ and the global record $M$
    \WHILE{the rollout limits are not reached}
      \IF{an Execution proposal $(y,e)$ is pending}
        \STATE Reflection either generates \tool{approve\_commit}$(p)$ for an available proposal $p$ or \tool{navigate}$(v,i)$
        \IF{Reflection approves}
          \RETURN the answer stored at $p$, $\mathcal{T}$
        \ENDIF
      \ELSE
        \STATE Reflection generates \tool{navigate}$(v,i)$ from the global record $M$
      \ENDIF
      \STATE Restore the path to $v$ and start Execution with instruction $i$
      \WHILE{the current Execution segment is active}
        \STATE Execution uses ordinary domain tools
        \IF{Execution generates \tool{checkpoint}$(s)$, or the harness requires one when the context is full}
          \STATE Save a child of $v$ with card $s$; append the result to $M$; break
        \ELSIF{Execution generates \tool{commit}$(y,e)$}
          \STATE Save proposal $(y,e)$ as a child of $v$; append it to $M$; break
        \ENDIF
      \ENDWHILE
    \ENDWHILE
    \RETURN no approved answer and the reason the rollout stopped
\end{algorithmic}
\end{algorithm}

\subsection{Context Layouts}
\label{app:context-layouts}

The harness builds each role's context from fixed templates. We show the templates with placeholders in braces and shorten long fixed instructions. Appendix~\ref{app:prompts} reproduces the system prompts.

\paragraph{Reflection.} Reflection runs as one conversation in which the model calls tools over many turns. Each \tool{navigate} call returns only when the Execution segment it starts has ended. Its result describes the new node and includes the node's card.

\begin{systemprompt}
[system]     {Reflection system prompt}
[user]       {task}  {navigation example}  {initial tree: root}
[assistant]  navigate(node="root", instruction="{objective 1}")
[tool]       [EXECUTION CHECKPOINT]
             node: n1
             parent: root
             path: root -> n1
             {deep research: card: {card}}
             {SWE: report: {card}  trusted_receipt: {harness summary: diffstat, changed files, test runs}}
             valid_navigation_nodes: [...]
             available_reflection_actions: ["navigate"]
             [/EXECUTION CHECKPOINT]
[assistant]  navigate(node="n1", instruction="{objective 2}")
[tool]       [EXECUTION COMMIT PROPOSAL]
             node: n2
             parent: n1
             path: root -> n1 -> n2
             {deep research: answer, explanation, supporting_docids}
             {SWE: report: {completion message}  trusted_receipt: {...}}
             available_reflection_actions: ["approve_commit", "navigate"]
             [/EXECUTION COMMIT PROPOSAL]
[assistant]  approve_commit(node="n2")          -> the search ends
\end{systemprompt}

In SWE, each result also gives the number of turns and navigations left, says whether the node's raw context is still saved, and briefly suggests where to navigate next.

\paragraph{Execution in deep research.} Each segment starts from a new context that holds the cards along the path and the current instruction, but no earlier instructions.

\begin{systemprompt}
[system]     {Execution system prompt}
[user]       {question}
[user]       [Resumed checkpoint n1]
             === ACTIVE PATH CARDS (root to current) ===
             root: {card}
             n1: {card}
             {available tools: search, get_document, checkpoint, commit}
             <EXECUTION_TASK>
             {instruction}
             </EXECUTION_TASK>
[assistant]  {tool calls and results, ending with checkpoint(summary) or commit(answer, explanation)}
\end{systemprompt}

\paragraph{Execution in SWE, with the full saved context.} The harness restores the raw context saved at the selected node, which holds every earlier segment on the path exactly as the model generated it. It then adds one user message that opens the new segment. In the layout below, only these opening messages come from templates, for example ``[Cursor restored to n1] \ldots''.

\begin{systemprompt}
[system]     {Execution system prompt}
[user]       {issue and repository location}
[user]       [Cursor restored to root] ... <EXECUTION_TASK> {instruction 1} </EXECUTION_TASK> ...
[assistant]  {reasoning} {tool call, e.g. execute_bash(...)}
[tool]       {tool result}
             ...
[assistant]  checkpoint(summary="{card of n1}")
[tool]       Checkpoint n1 saved; control returns to REFLECTION.
[user]       [Cursor restored to n1] ... <EXECUTION_TASK> {instruction 2} </EXECUTION_TASK> ...
[assistant]  ...                                      (segment 2, as generated)
[user]       [Cursor restored to n2]
             The saved repository and raw Execution path are active.
             turns_remaining: {k}
             <EXECUTION_TASK>
             {instruction 3}
             </EXECUTION_TASK>
             <SEGMENT_BOUNDARY>
             {stopping rules: checkpoint once the objective is resolved or blocked}
             </SEGMENT_BOUNDARY>
\end{systemprompt}

\paragraph{Execution in SWE, with a shortened context.} When a node is saved because the context is full, the harness replaces its saved context with a single message that holds the cards along the path and a shortened diff of the repository. Earlier nodes keep their raw context.

\begin{systemprompt}
[system]     {Execution system prompt}
[user]       {issue and repository location}
[user]       [Compacted checkpoint n3]
             The earlier raw Execution transcript on this path was compacted. The checkpoint
             cards below are the complete record of prior work on this path, oldest first.
             [n1] objective: {instruction, at most 150 tokens}
             card: {card, at most 600 tokens}
             [n2] ...
             [n3] (this checkpoint) objective: {...}
             card: {...}
             [Repository diff relative to the original checkout]
             {diffstat, changed files, and a patch prefix, together at most 2,048 tokens}
             The exact repository state of n3 is restored; ...
[user]       [Cursor restored to n3]
             The exact saved repository is active. Earlier raw tool output was compacted;
             rely on the retained path cards and re-inspect anything decisive.
             turns_remaining: {k}
             <EXECUTION_TASK> {instruction} </EXECUTION_TASK>
             <SEGMENT_BOUNDARY> {stopping rules} </SEGMENT_BOUNDARY>
\end{systemprompt}

\section{Training Details}
\label{app:training-details}

\subsection{Causal Block Packing}
\label{app:block-packing}

Each training row has a prompt part and a response part of fixed length, and the loss applies only to tokens in the response part. A causal block starts with restored context and continues with the tokens the model generated in that segment. We put as much of the restored context as fits into the prompt part, and everything from the first generated token onward into the response part. Restored context that does not fit goes into the response part but is left out of the loss, so the model still sees it but is not trained on it. If the response part is too long, we split the block into several rows, and we split only between model calls. Every token is therefore trained on exactly the context it was generated from.

\subsection{Building PPO Minibatches}

Each causal block becomes one training row, or several rows if it is too long (Appendix~\ref{app:block-packing}). All rows of a rollout share its advantage. The group mean and standard deviation are computed over rollouts, so a rollout counts once however many rows it produces. The rollouts of one prompt form a prompt group. All rows from a prompt group go into the same PPO minibatch, so we count minibatch size in prompt groups, not rows. We compute the weights of Equation~\eqref{eq:treeflex-objective} from token counts over the whole minibatch, before the minibatch is split across GPUs and microbatches. Padding rows therefore get zero weight, and the split does not change the token mean or the rollout mean. The block mean weights every row equally, so a block split into several rows counts once per row.

\subsection{Domain Rewards}
\label{app:reward}

We use the notation of Section~\ref{sec:training}. Rollout $i$ receives a terminal reward $R_i$, and token $t$ in causal block $b$ of that rollout receives a process penalty $Q_{i,b,t}$. Here $\mathbf{1}[\cdot]$ is $1$ if its condition holds and $0$ otherwise.

\paragraph{Terminal reward.} For BrowseComp-Plus, $R_i$ is 1 if the submitted answer is judged correct and 0 otherwise. Flat \react{} submits its answer by calling \tool{commit}, \method{} submits the proposal that Reflection approves, and \foldagent{} submits when its main agent calls \tool{finish}.

\begin{equation}
R_i = \mathbf{1}\bigl[\text{rollout $i$ submits a valid answer}\ \land\ \text{the judge marks it correct}\bigr].
\end{equation}

For SWE, the terminal reward comes from the official grading of one candidate state of the repository.

\begin{equation}
R_i = \mathbf{1}\bigl[\text{the candidate's grade is valid}
\ \land\ \text{all fail-to-pass and pass-to-pass tests pass}\bigr].
\end{equation}

A candidate receives $R_i=0$ if its grade is invalid or if it does not pass all the tests. During training, the candidate for \method{} is the proposal that Reflection approved. A rollout with no approved proposal receives $R_i=0$. For \foldagent{} with its released prompt, we grade the final workspace whether or not the main agent called \tool{finish}. The step-wise variant follows the same rule as \method{}. A rollout receives $R_i=0$ unless the main agent calls \tool{finish} (Section~\ref{sec:results}). At evaluation, \method{} and \foldagent{} choose the candidate by the rule of Section~\ref{sec:exp-setup}.

\paragraph{Process penalties.} On BrowseComp-Plus, \method{} uses two of the three process penalties of \foldagent{}~\citep{sun2025foldagent}, the failure penalty and the out-of-scope penalty. The ablations of Section~\ref{sec:ablations} add a Reflection finalization penalty. On SWE, we use only the failure penalty. Every penalty applies only to successful rollouts, those with $R_i=1$.

\begin{equation}
Q_{i,b,t} =
\begin{cases}
-1, & \text{if $R_i=1$ and $t$ belongs to a turn whose tool call failed},\\
-0.2, & \text{if $R_i=1$ and $t$ belongs to an out-of-scope retrieval action},\\
-1, & \text{if $R_i=1$ and $t$ belongs to an answer-requesting Reflection instruction (ablations)},\\
0, & \text{otherwise}.
\end{cases}
\end{equation}

When more than one penalty applies to a token, we use the most negative one. The failure penalty applies only to the failed generated action. In BrowseComp-Plus, the out-of-scope penalty applies only to generated retrieval actions labelled out of scope, and never to checkpoint or proposal turns. In the ablations, the finalization penalty applies to a Reflection instruction that asks Execution to produce the answer instead of finding new evidence. A fixed text pattern or the OOS judge flags such instructions. Three kinds of instruction are exempt. They are requests to write up a card, verification requests such as ``finalize whether X holds by checking Y'', and the final instruction Reflection gives when its navigations or turns run out, since that last segment must end with a proposal. We leave out of training any rollout whose correctness judgment is ambiguous. We give no positive reward for nodes, navigation targets, or tree shape. Appendix~\ref{app:bcp-judges} gives the judges, and Appendix~\ref{app:hyperparams} gives optimizer, evaluator, and sampling settings.

\section{Deep-Research Setup}
\label{app:bcp-setup}

\subsection{Data Split}
\label{app:bcp-data}

The 64 validation questions are held out from the 680 training questions of \citet{sun2025foldagent}. Their BrowseComp-Plus query IDs are listed below.

{\small
\begin{quote}
3, 6, 15, 85, 86, 111, 152, 154, 179, 181, 210, 236, 283, 303, 310, 333, 391, 416, 449, 481, 501, 503, 524, 532, 535, 555, 572, 579, 583, 596, 624, 665, 692, 719, 742, 760, 770, 775, 781, 896, 920, 952, 959, 1000, 1010, 1019, 1021, 1029, 1040, 1049, 1053, 1068, 1094, 1098, 1126, 1163, 1174, 1179, 1218, 1234, 1240, 1242, 1247, 1250
\end{quote}
}

\subsection{Context Limits}
\label{app:context-limits}

The main Qwen3-4B RL comparison limits the active context to 32,768 tokens. In all runs, each model call generates at most 3,072 tokens and a rollout has at most 100 turns. All runs also cut retrieved text the same way. A fresh result keeps 512 tokens, a previously seen result keeps 128, a document keeps at most 4,096, and at most ten retrieved items are kept. Retrieved text counts toward the context length, but not toward the generated tokens reported in Table~\ref{tab:main-results}. Total generated tokens are not capped.

\paragraph{TreeFlex.} Before each Execution turn, the harness estimates whether the next action and its tool result would still fit in the 32,768-token active context. If they would not fit, the model must save a checkpoint before it does any more work on the task. Reflection then chooses the next node and instruction. The model window of 49,152 tokens is larger than the 32,768-token active context, which leaves room for the prompt templates and the closing message. The RL run sets no separate limit on turns per segment.

\paragraph{FoldAgent.} The reported \foldagent{} run caps each branch's active context at 32,768 tokens and allows at most ten branches. Its 49,152-token model window keeps 3,072 tokens for closing a branch and leaves the rest for prompt templates and observations. When a branch finishes, the branch's report replaces the branch in the main context. Apart from these replacements, the main context only grows.

\paragraph{Flat \react{}.} The 32k flat \react{} run keeps all turns in one context. That context is limited to 32,768 tokens, inside the same 49,152-token model window that \foldagent{} and \method{} use. We set the difficulty bins of Figure~\ref{fig:difficulty} using a separate flat \react{} run with a 40,960-token window. The 254k diagnostic run reserves 8,192 tokens for the initial prompt and lets the rest of the context grow to 253,952 tokens in a 262,144-token model window. We take 16 audited rollouts from each of the last ten training steps before the reported step, and measure the largest context each rollout reaches, counting the initial prompt. The median is 25.5k tokens for the 32k run and 29.1k for the 254k run. The 254k run has a long tail, with 12\% of rollouts above 64k tokens, a 90th percentile of 87.1k, and a maximum of 254.6k. The 32k run stays near its limit, with a 90th percentile of 33.3k and no rollout above 64k.

\subsection{Reward Judges}
\label{app:bcp-judges}

\paragraph{Precision-first outcome judge.} All BrowseComp-Plus methods are scored by the same fixed judge. The released \foldagent{} implementation normalizes parenthetical qualifiers, years, function words, token order, title prefixes, and some first names before invoking its language-model judge~\citep{sun2025foldagent}. This normalization can let a partial answer count as correct, so we add a second check that tends to reject wrong answers.

An answer is accepted without the language-model judge only if it exactly matches the gold answer, a recorded alias for that question, or the same number written plainly. Our normalization rules are conservative. They can flag two answers as possibly equivalent, but they cannot mark an answer correct on their own. Fixed rules reject an answer if its numbers conflict with the gold answer, if it changes a sign or a percentage, if it adds a qualifier to a title, or if it adds a bibliographic suffix that the gold answer does not support.

Every other answer is judged once by Qwen3-32B under a fixed prompt. If Qwen3-32B's verdict disagrees with the fixed rules, or if the answer looks suspiciously close to the gold answer, we sample three more verdicts. We use these verdicts only if all three agree. If the three verdicts disagree, the answer is ambiguous, and we leave the rollout out of the policy loss and the GRPO group statistics. With this check, the judge accepts fewer wrong answers but also rejects some correct ones.

\paragraph{Per-action out-of-scope judge.} \foldagent{} decides whether a branch stayed in scope by comparing the branch's assignment with the report it returns~\citep{sun2025foldagent}. For \method{}'s Execution segments, we judge each retrieval action instead, since a report can leave out work or credit work to the wrong step. The judge sees only the original question, the card of the selected node, the assigned instruction, and up to eight recent instructions given at that node. For each \tool{search} or \tool{get\_document} call, it also sees the truncated arguments, information about the results, and whether the call succeeded. The judge does not see the gold answer, the reward, or whether the final answer was correct.

{\emergencystretch=1em The Qwen3-32B judge also labels each instruction as focused, broad, redundant, finalization, or uncertain. The instruction labels do not change the OOS penalty. The ablations use the finalization label to apply the Reflection finalization penalty. Each action is labeled in scope, necessary verification, redundant, out of scope, or uncertain. Only an out-of-scope action is penalized, with $Q=-0.2$ on the turn that made the call. The judge's output must label every action exactly once. If the judge's output is malformed, or the request to the judge fails or times out, we apply no out-of-scope penalty in that segment. Appendix~\ref{app:oos-prompt} gives the exact prompt.\par}

\subsection{FoldAgent Baseline}

The \foldagent{} authors released an open-source reimplementation of their method, which may differ from the code behind their paper. We run it in the same training and evaluation code as the other methods, with its prompts and tool definitions unchanged, so all methods share the retriever, the answer judge, and the trainer. All reported \foldagent{} results use this implementation.

\subsection{Hyperparameters}
\label{app:hyperparams}

Table~\ref{tab:hyperparams} lists the configuration of the four RL runs reported in Table~\ref{tab:main-results}. Each ablation run in Table~\ref{tab:lambda-ablation} differs from the \method{} run only in the settings it studies, with one exception. The two runs with the finalization penalty also use a 40,960-token model window instead of 49,152 tokens, and keep the 32k active-context limit.

\begin{table}[h]
  \centering
  \caption{\textbf{Configuration of the reported BrowseComp-Plus runs.} A value that spans all columns applies to every run. Sampling rows are temperature / top-$p$ / top-$k$ / min-$p$. Context values are in tokens.}
  \label{tab:hyperparams}
  \scriptsize
  \begin{tabular}{lcccc}
    \toprule
    \textbf{Setting} & \textbf{\method{}} & \textbf{\foldagent{}} & \textbf{Flat \react{} (32k)}
      & \textbf{Flat \react{} (254k)} \\
    \midrule
    \multicolumn{5}{l}{\textbf{Model and data}} \\
    Base model & \multicolumn{4}{c}{Qwen3-4B-Instruct-2507} \\
    Train / validation / test questions & \multicolumn{4}{c}{616 / 64 / 150} \\
    Machines $\times$ GPUs per machine & \multicolumn{4}{c}{$2\times8$} \\
    \midrule
    \multicolumn{5}{l}{\textbf{Optimization}} \\
    Training steps & \multicolumn{4}{c}{50} \\
    Prompts per step / samples per prompt & \multicolumn{4}{c}{32 / 8} \\
    PPO minibatch (prompt groups) & \multicolumn{4}{c}{16} \\
    Learning rate (no warmup) & \multicolumn{4}{c}{$1\times10^{-6}$} \\
    Clip range (low / high), dual-clip constant & \multicolumn{4}{c}{0.2 / 0.28, 3.0} \\
    KL penalty & \multicolumn{4}{c}{none} \\
    Policy optimization & tree-aware GRPO (\S\ref{sec:training}) & FoldGRPO & GRPO & GRPO \\
    Advantage std normalization & \multicolumn{4}{c}{yes} \\
    Loss aggregation & rollout mean, $\lambda=0$ & \multicolumn{3}{c}{token mean} \\
    Filter zero-variance groups & \multicolumn{4}{c}{no} \\
    Training sampling & \multicolumn{4}{c}{1.0 / 1.0 / $-1$ / 0} \\
    \midrule
    \multicolumn{5}{l}{\textbf{Context and interaction limits}} \\
    Active-context limit & 32,768 & 32,768 per branch & 32,768 & 253,952 \\
    Model window & 49,152 & 49,152 & 49,152 & 262,144 \\
    Prompt / response columns & 8,192 / 40,960 & 8,192 / 40,960 & 8,192 / 40,960 & 8,192 / 253,952 \\
    Max new tokens per call & \multicolumn{4}{c}{3,072} \\
    Max model turns & \multicolumn{4}{c}{100} \\
    Max navigations or branches & 10 navigations & 10 branches & -- & -- \\
    Context-pressure rule & checkpoint when full & off & off & off \\
    Retrieved document cap / snippet cap & \multicolumn{4}{c}{4,096 / 512} \\
    \midrule
    \multicolumn{5}{l}{\textbf{Rewards}} \\
    Terminal reward & \multicolumn{4}{c}{binary judge correctness} \\
    Failure / out-of-scope penalty & $-1$ / $-0.2$ & $-1$ / $-0.2$ & -- & -- \\
    Unfolded-context penalty & -- & $-1$ & -- & -- \\
    Answer judge & \multicolumn{4}{c}{Qwen3-32B, 1 vote; 3-vote adjudication on disagreement; sampling 0.7 / 0.8 / 20} \\
    \midrule
    \multicolumn{5}{l}{\textbf{Evaluation and selection}} \\
    Validation / test sampling & \multicolumn{4}{c}{0.7 / 0.8 / 20 / 0} \\
    Validation samples / cadence & \multicolumn{4}{c}{3 per question / every 5 steps} \\
    Test samples per question & \multicolumn{4}{c}{8} \\
    Model selection & \multicolumn{4}{c}{highest validation Mean@3 accuracy} \\
    Selected step & 45 & 40 & 45 & 40 \\
    \bottomrule
  \end{tabular}
\end{table}

\subsection{Training Dynamics}
\label{app:training-dynamics}

\begin{figure}[t]
  \centering
  \includegraphics[width=0.68\textwidth]{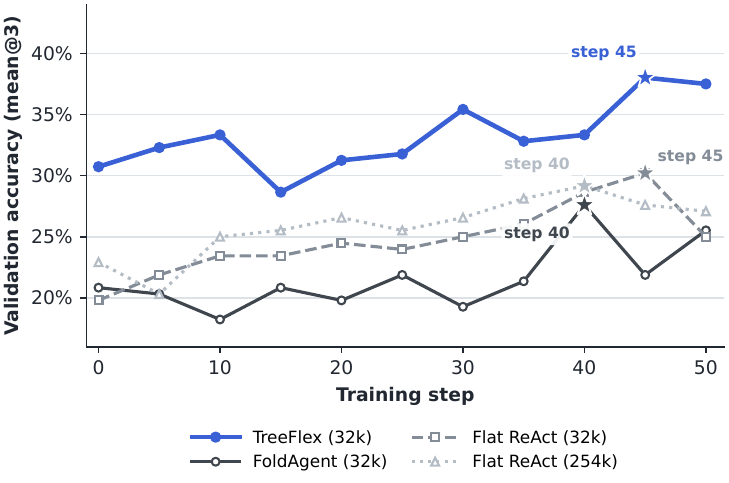}
  \caption{\textbf{TreeFlex leads on validation throughout training.} Curves show unsmoothed Mean@3 on the fixed 64-question validation split every five steps. Stars mark the steps selected for test evaluation. The plot does not show test accuracy.}
  \label{fig:validation-dynamics}
\end{figure}

Figure~\ref{fig:validation-dynamics} tracks validation Mean@3 during BrowseComp-Plus training. Tree-aware GRPO raises \method{}'s validation Mean@3 from 30.7\% at step 0 to 38.0\% at the selected step, and \method{} leads all three baseline runs at every evaluated step.

\FloatBarrier

\section{SWE Setup}
\label{app:swe-setup}

\subsection{Training and Evaluation Tasks}
\label{app:data-splits}

We use a 600-task SWE-Gym subset stratified by repository and by the difficulty of the reference solution, and split it into 500 training and 100 validation tasks. For task $j$, let $L_j$ be the number of changed lines in the reference solution, $F_j$ the number of changed files, and $T_j$ the number of fail-to-pass tests. We compute

\begin{equation}
d_j = \log(1+L_j) + 0.5\log(1+F_j) + 0.25\log(1+T_j),
\end{equation}

and sort the tasks of each repository by $d_j$ into easy, medium, and hard thirds. We then pick the 100 validation tasks so that each group of repository and difficulty third contributes in proportion to its size, rounding the shares with the largest-remainder method. We use the validation tasks only to choose which saved training step to report, and we run three rollouts per task. The difficulty score is used only to choose the validation tasks and is hidden from the agent.

\paragraph{Training subsets.} GRPO learns nothing from the task reward when all rollouts of a task fail or all succeed. We therefore keep the training tasks on which a method's earlier training rollouts both succeeded and failed. For \method{}, a rollout counts as successful only if it resolves the task with an approved proposal. The filter keeps 217 tasks for \foldagent{} and 191 for \method{}. Of these, 158 are in both sets and 250 are in at least one. \method{} and the step-wise \foldagent{} variant train on the union, and \foldagent{} trains on its own 217 tasks, because the other 33 never gave it mixed outcomes. We pick these subsets using only training tasks and their training rollouts.

\subsection{Repository Restoration}
\label{app:swe-state}

When a node is saved, the harness records how the agent has changed the repository since the task's base commit. Saving does not change any branch or commit that the agent can see. Restoring a node returns the repository to this saved state and starts a new shell. Shell variables and running processes do not carry over to the restored node.

Each rollout runs in its own writable copy of the task's container image. The model has OpenHands-compatible \tool{execute\_bash}, \tool{str\_replace\_editor}, \tool{think}, and \tool{finish} tools. In \method{}, Git commands that would change branches or commits are blocked, but the read-only commands status, diff, log, show, and blame remain available. The model never sees the reference patch, the test patch, or the official grader's metadata.

\subsection{Comparison with FoldAgent}
\label{app:swe-alignment}

\paragraph{Method-specific settings.}
\foldagent{} uses its released SWE prompt and protocol, and the sampling settings given below. On SWE, \method{} uses two additional settings. After a checkpoint forced by the context limit, the next Execution segment on that path starts from the compacted context of Appendix~\ref{app:context-layouts}. Each time a role takes over, the harness tells it how many turns remain. Reflection is also told how many navigations remain.

\paragraph{Matched model, environment, and limits.}
The reported rows use the post-trained \texttt{Qwen/Qwen3.5-9B} model with thinking enabled.

Both methods run the model with a context window of 57,344 tokens. The initial prompt may take up to 16,384 tokens, and the context may then grow by at most 32,768 tokens before the method must compact it or stop. The rest of the window leaves room for the next call. A rollout may use at most 200 model turns and 900 seconds per generation. At evaluation, it may run for at most 90 minutes in total. Tool outputs are cut to 7,680 tokens and 500 lines.  Forced checkpoints and branch reports are capped at 4,096 tokens.

Both methods use the same container images and offer the same kinds of tools. In both, shell commands time out after 300 seconds, and the same lenient parser reads tool calls written after the model's thinking. Both use the official SWE-bench grader and its timeouts. All evaluations use temperature 0.6, top-$p$ 0.95, top-$k$ 20, and min-$p$ 0.

\paragraph{Training protocol.} Both methods use PPO minibatches of 16 tasks and normalize advantages by the group standard deviation. Every run saves its model every five steps. Ties in validation Mean@3 go to the earlier step, and the selection picks step 25 for \method{} and step 10 for both \foldagent{} variants.
During training, both sample with temperature 1, top-$p$ 1, no top-$k$ cutoff (top-$k=-1$), and min-$p$ 0. \method{} and \foldagent{} with its released prompt keep the failure penalty and turn off the out-of-scope and unfolded-context penalties. The out-of-scope and unfolded-context penalties assume that the agent branches. For a successful rollout that never branches, the released \foldagent{} code penalizes every turn and by default sets the reward to zero. \foldagent{} branches in under 1\% of SWE rollouts, so keeping the two penalties would remove almost all of its learning signal.

Besides their training tasks, the runs differ in three respects. \method{} uses the rollout mean of Equation~\eqref{eq:treeflex-objective} with $\lambda=0$, whereas \foldagent{} uses the token mean of FoldGRPO~\citep{sun2025foldagent}. Their training rewards differ as described in Appendix~\ref{app:reward}. \method{} caps each training rollout at 45 minutes and \foldagent{} at 90. Evaluation uses the same 90-minute limit and grading rule for both.

The step-wise delegation variant of \foldagent{} (Table~\ref{tab:swebench}) also adds FoldGRPO's unfolded-context penalty.

\paragraph{Mechanism-specific differences.} \foldagent{} keeps one main context that only grows, and may launch up to ten short-lived branches that all edit the same workspace. Each branch returns a report, or its last message if it writes no report. A branch cannot restore an earlier repository state.

In \method{}, Execution may call only \tool{finish} in the last three turns of the rollout, counting the turns of both roles. \foldagent{} has no rule that limits its last turns to \tool{finish}, and its rollout stops when it reaches the turn or context limit.

The tool interfaces also differ. In \method{}, each \tool{execute\_bash(command)} call runs in a fresh shell, editor arguments are checked strictly, and \tool{finish} requires a message. \foldagent{} keeps one shell open across calls, and its shell tool takes \tool{is\_input} and \tool{timeout} arguments. It also keeps its original editor, and its \tool{finish} message is optional.

\section{Additional Results}
\label{app:results}

\subsection{Paired Comparisons}
\label{app:paired}

Table~\ref{tab:paired} lists the BrowseComp-Plus paired comparisons discussed in Section~\ref{sec:results}. Each difference is \method{}'s score minus the compared system's score on the same 150 test questions, in percentage points. The 95\% intervals come from a paired bootstrap that resamples the questions 10,000 times. These intervals capture variation from the choice of questions, but not variation across training seeds. For the voting rows, we compute majority-vote accuracy averaged exactly over every subset of $k$ of the eight samples.

\begin{table}[h]
  \centering
  \caption{\textbf{Paired differences on BrowseComp-Plus.} Bold marks intervals that exclude zero.}
  \label{tab:paired}
  \small
  \begin{tabular}{llrc}
    \toprule
    \textbf{Metric} & \textbf{Comparison} & \textbf{Difference} & \textbf{95\% interval} \\
    \midrule
    Mean@8 & vs.\ flat \react{} & $+7.33$ & \textbf{[3.50, 11.25]} \\
             & vs.\ \foldagent{} & $+6.00$ & \textbf{[1.67, 10.42]} \\
    Pass@8 & vs.\ flat \react{} & $+10.00$ & \textbf{[4.00, 16.00]} \\
           & vs.\ \foldagent{} & $+10.67$ & \textbf{[4.67, 16.67]} \\
    \midrule
    Vote@8 & vs.\ flat \react{} vote@8 & $+9.08$ & \textbf{[3.36, 14.97]} \\
           & vs.\ \foldagent{} vote@8 & $+4.64$ & [$-0.83$, 10.33] \\
    \method{} $k{=}1$ & vs.\ flat \react{} vote@5 (27.0 calls) & $+4.82$ & \textbf{[0.58, 9.11]} \\
                      & vs.\ flat \react{} vote@6 (32.4 calls) & $+4.53$ & \textbf{[0.19, 8.90]} \\
                      & vs.\ \foldagent{} vote@2 (25.7 calls) & $+6.00$ & \textbf{[1.67, 10.42]} \\
                      & vs.\ \foldagent{} vote@3 (38.5 calls) & $+3.52$ & [$-0.90$, 8.00] \\
    \midrule
    Mean@8, easy & vs.\ flat \react{} & $+2.16$ & [$-2.37$, 7.11] \\
    Mean@8, medium & vs.\ flat \react{} & $+13.50$ & [$-1.00$, 27.00] \\
    Mean@8, hard & vs.\ flat \react{} & $+9.51$ & \textbf{[4.29, 15.30]} \\
    Mean@8, easy & vs.\ \foldagent{} & $+7.54$ & \textbf{[1.51, 14.22]} \\
    Mean@8, medium & vs.\ \foldagent{} & $+5.00$ & [$-11.00$, 20.00] \\
    Mean@8, hard & vs.\ \foldagent{} & $+5.04$ & [0.00, 10.26] \\
    \midrule
    Mean@8 & vs.\ block mean, $\lambda=0$ & $+0.42$ & [$-2.25$, 3.00] \\
    Pass@8 & vs.\ block mean, $\lambda=0$ & $+0.67$ & [$-4.00$, 5.33] \\
    Mean@8 & vs.\ rollout mean, $\lambda=0.3$ & $+3.58$ & \textbf{[1.08, 6.17]} \\
    Pass@8 & vs.\ rollout mean, $\lambda=0.3$ & $+0.00$ & [$-5.33$, 5.33] \\
    Mean@8 & vs.\ token mean, $\lambda=0.3$ & $+1.17$ & [$-1.33$, 3.67] \\
    Pass@8 & vs.\ token mean, $\lambda=0.3$ & $+0.00$ & [$-4.00$, 4.00] \\
    Mean@8 & vs.\ token mean, $\lambda=0$ & $+2.17$ & [$-0.33$, 4.75] \\
    Pass@8 & vs.\ token mean, $\lambda=0$ & $+0.67$ & [$-3.33$, 5.33] \\
    Mean@8 & vs.\ token mean, $\lambda=0$, finalization penalty & $+2.42$ & [$-0.17$, 5.08] \\
    Pass@8 & vs.\ token mean, $\lambda=0$, finalization penalty & $+0.00$ & [$-4.67$, 4.67] \\
    Mean@8 & vs.\ token mean, $\lambda=0.3$, finalization penalty & $+4.67$ & \textbf{[2.42, 7.00]} \\
    Pass@8 & vs.\ token mean, $\lambda=0.3$, finalization penalty & $+4.00$ & [0.00, 8.00] \\
    \bottomrule
  \end{tabular}
\end{table}

Table~\ref{tab:paired-swe} lists paired differences in Mean@3 on the same 500 SWE-bench Verified tasks. The intervals use the same paired bootstrap, resampling tasks instead of questions.

\begin{table}[h]
  \centering
  \caption{\textbf{Paired differences on SWE-bench Verified (Mean@3).} Bold marks intervals that exclude zero.}
  \label{tab:paired-swe}
  \small
  \begin{tabular}{lrc}
    \toprule
    \textbf{Comparison} & \textbf{Difference} & \textbf{95\% interval} \\
    \midrule
    \method{} vs.\ \foldagent{}, base models & $+19.47$ & \textbf{[16.20, 22.73]} \\
    \method{} vs.\ \foldagent{} step-wise, base models & $+1.60$ & [$-0.73$, 4.00] \\
    \method{} RL vs.\ \foldagent{} RL & $+11.60$ & \textbf{[9.00, 14.27]} \\
    \method{} RL vs.\ \foldagent{} step-wise RL & $+9.73$ & \textbf{[7.07, 12.40]} \\
    \midrule
    \method{} RL vs.\ its base model & $+3.80$ & \textbf{[1.40, 6.27]} \\
    \foldagent{} RL vs.\ its base model & $+11.67$ & \textbf{[9.00, 14.40]} \\
    \foldagent{} step-wise RL vs.\ its base model & $-4.33$ & \textbf{[$-6.67$, $-2.07$]} \\
    \midrule
    Rollout mean vs.\ token mean, RL & $+2.80$ & \textbf{[0.60, 5.00]} \\
    Token mean RL vs.\ base model & $+1.00$ & [$-1.33$, 3.33] \\
    \bottomrule
  \end{tabular}
\end{table}
\FloatBarrier

\subsection{SWE Outcomes by Submission}
\label{app:swe-finish}

Table~\ref{tab:swebench} grades one repository state per rollout (Section~\ref{sec:exp-setup}), whether or not the policy submitted it. Table~\ref{tab:swe-finish} adds a stricter score that counts a task as resolved only when the policy itself submitted the solution. For \foldagent{}, the policy submits when the main agent calls \tool{finish}. For \method{}, it submits when Reflection approves the proposal. \method{} submits most of its solutions, so the two scores are close. With its released prompt, most of \foldagent{}'s resolved rollouts come from grading a workspace it did not submit.

\begin{table}[h]
  \centering
  \caption{\textbf{Most of the tasks TreeFlex resolves are solutions the policy submitted.} Resolved is the Mean@3 of Table~\ref{tab:swebench}. Submitted and resolved counts a rollout only if the policy submitted the graded solution. Submitted is the share of rollouts that end with \tool{finish} or an approved proposal.}
  \label{tab:swe-finish}
  \small
  \setlength{\tabcolsep}{4.5pt}
  \begin{tabular}{llrrr}
    \toprule
    \textbf{System} & \textbf{RL} & \textbf{Resolved} & \textbf{Submitted and resolved} & \textbf{Submitted} \\
    \midrule
    \multirow{2}{*}{\foldagent{}} & none & 31.7 & 3.7 & 3.9 \\
      & FoldGRPO & 43.3 & 13.4 & 17.4 \\
    \addlinespace
    \multirow{2}{*}{\shortstack[l]{\foldagent{}, step-wise\\delegation prompt}} & none & 49.5 & 31.5 & 42.3 \\
      & FoldGRPO & 45.2 & 40.8 & 69.5 \\
    \addlinespace
    \multirow{2}{*}{\method{}} & none & 51.1 & 50.1 & 86.1 \\
      & tree-aware GRPO & 54.9 & 53.6 & 90.1 \\
    \bottomrule
  \end{tabular}
\end{table}

\subsection{Search Diversity and Evidence Coverage}
\label{app:evidence}

For the 4B RL models of Table~\ref{tab:main-results}, Table~\ref{tab:evidence} counts the distinct queries, opened documents, and evidence documents each method reaches over the eight test samples of a question. Search queries are compared after lowercasing and removing punctuation. A document counts as read only when the agent opens it with \tool{get\_document} or \tool{open\_page}. Search-result snippets do not count. A document opened several times counts once. Evidence documents are all documents that BrowseComp-Plus marks as evidence or as containing the answer~\citep{chen2025browsecompplus}. All values are averages over the 150 test questions.

\begin{table}[h]
  \centering
  \caption{\textbf{Repeated flat sampling revisits the same evidence.} ``Per rollout'' averages over the eight samples of a question. ``Across 8'' pools the eight samples. Repeated is the share of a question's queries that duplicate another of its queries. Evidence read is the share of a question's annotated evidence documents that were opened.}
  \label{tab:evidence}
  \small
  \begin{tabular}{lrrrrrrr}
    \toprule
    & \multicolumn{3}{c}{\textbf{Search queries}} & \multicolumn{2}{c}{\textbf{Distinct opened documents}}
      & \multicolumn{2}{c}{\textbf{Evidence read}} \\
    \cmidrule(lr){2-4}\cmidrule(lr){5-6}\cmidrule(lr){7-8}
    \textbf{Method} & \textbf{Per rollout} & \textbf{Distinct, across 8} & \textbf{Repeated}
      & \textbf{Per rollout} & \textbf{Distinct, across 8} & \textbf{Per rollout} & \textbf{Across 8} \\
    \midrule
    Flat \react{} & 3.4 & 21.3 & 21\% & 2.0 & 5.3 & 19\% & 34\% \\
    \foldagent{} & 9.3 & 68.0 & 10\% & 2.7 & 8.7 & 26\% & 48\% \\
    \method{} & 20.7 & 153.2 & 6\% & 5.2 & 16.0 & 34\% & 59\% \\
    \bottomrule
  \end{tabular}
\end{table}

\subsection{Case Study: Evidence-Directed Backtracking}
\label{app:case-study}

\begin{figure}[!t]
  \centering
  \includegraphics[width=\textwidth]{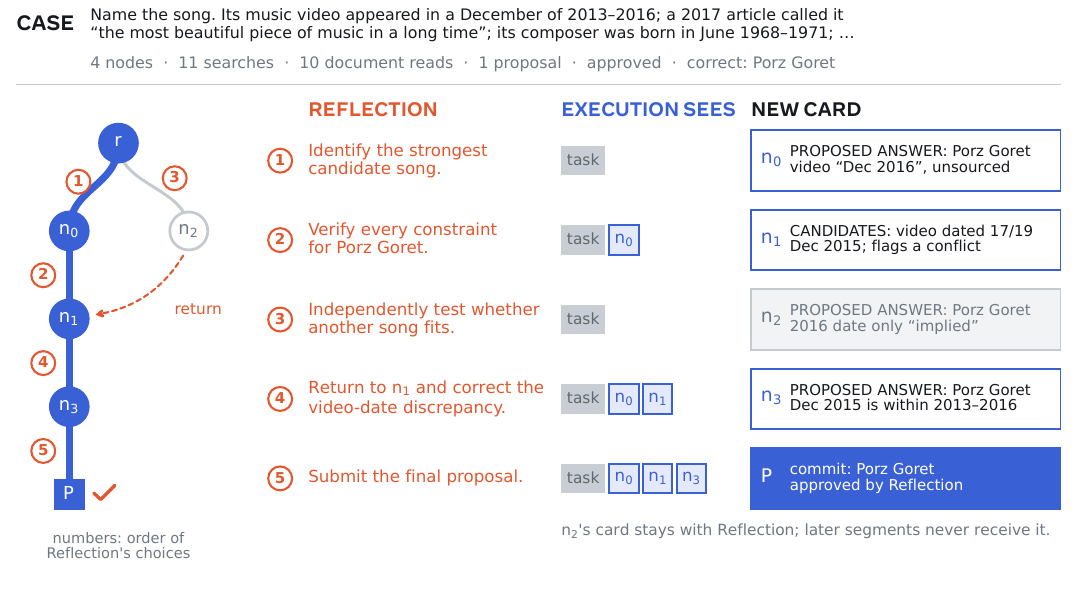}
  \caption{\textbf{TreeFlex restarts from root and then returns to an older node.} Badges mark Reflection's choices, and the dashed arrow shows the return from $n_2$ to $n_1$. Node $n_2$ starts a new branch from the root and finds no rival song. Node $n_3$ then resolves the remaining date conflict, and Reflection approves the answer.}
  \label{fig:case-study}
\end{figure}

Figure~\ref{fig:case-study} shows a successful Qwen3-4B rollout from training step 9.

The question asks for a song from clues about a video date, a 2017 article, a related album, and the composer's birth date. The benchmark answer is Yann Tiersen's ``Porz Goret''. The figure reconstructs every Reflection decision and node. We paraphrase the instructions and cards but keep the tree shape and the order of events.

The first segment proposes ``Porz Goret'' at $n_0$ but dates the video to December 2016 without evidence. Node $n_1$ finds press releases from December 2015 but misreads them as a conflict. Reflection then starts a new branch from the root. Node $n_2$ finds no rival song, so Reflection returns to $n_1$. Execution rechecks the sources, recognizes that 2015 lies within the requested range, and saves $n_3$. The next segment submits the answer, which Reflection approves.

The trace shows three behaviors. First, the model saves a checkpoint to note an open question and later turns that question into the instruction for a new segment. Second, it tries a sibling branch and then resumes the line that holds the evidence. The Execution segment on that line does not see the sibling's card. Third, during Execution it corrects a flawed instruction using the documents it reads. Throughout, the harness only restores the states that the model selects.

\FloatBarrier

\section{System Prompts}
\label{app:prompts}

We reproduce the system prompts verbatim.

\subsection{BrowseComp-Plus Reflection}

\begin{systemprompt}
You are REFLECTION, the persistent navigator for a hard multi-hop research task.

EXECUTION searches, reads documents, writes checkpoint cards, and drafts commit proposals. You do
none of those things. You select one saved Execution state for each bounded segment, give it one
focused instruction, and approve or reject its final proposal. approve_commit(node) submits the
proposal at that exact node as written.

Your conversation is a persistent chronological ledger. Each checkpoint result adds its node,
parent, path, cause, card, and valid navigation handles. Instructions are plans, not evidence.
Execution sees only the ordered cards on the path selected by navigate; off-path cards remain visible
only in this ledger.

Call navigate(node, instruction). Choose the node by premise fit: continue from the latest useful
node when its premises are needed, choose an earlier node to discard a later assumption, and choose
root for a genuinely independent line. Give one objective naming the decisive gap or hypothesis,
the evidence that would resolve it, and what to preserve. Do not bundle separate tasks.

Open another segment only for unsupported fields, conflicts, or evidence-backed alternatives.
After retrieval fails, change the test; failure to retrieve is not disproof. Never use navigate to
request submission, commit, finalization, proposal, or restatement; Execution decides when to
commit. Without a proposal, assign one decisive check. Do not repeat resolved work or
request generic auditing.

An Execution commit is a resumable proposal retained after later navigation. Call
approve_commit(node) only when that node's answer supplies every requested field and agrees with
the checkpoint ledger. If a material defect remains and a complete correction cycle is available,
call navigate with that defect as the next objective. Otherwise withhold approval with one concise
reason. You cannot edit a proposal.

If final_execution_only is true, use the last Execution segment to obtain the best grounded
proposal. Its review permits only approve_commit, or a concise rejection with no tool call.

TOOL-CALL CONTRACT:
- Without a proposal, issue exactly one structured navigate tool call and no prose.
- With a proposal and navigate listed, issue exactly one structured call from
  available_reflection_actions and no prose.
- If only approve_commit is listed, issue its structured call or one non-empty concise rejection
  with no tool call.
- Use the function-call channel defined by the available tool schema. Ordinary prose naming an
  action is not a tool call.
- Empty output, multiple calls, and unlisted tools are invalid. The latest action list supersedes
earlier ones.
\end{systemprompt}

\subsection{BrowseComp-Plus Execution}

\begin{systemprompt}
You are a deep-research agent answering a hard, multi-hop question by iterative retrieval over a frozen corpus. You reason step by step and use tools ONE AT A TIME.

DOMAIN TOOLS:
- search(query): top document snippets with docids. Write descriptive, natural-language queries (the retriever is embedding-based — full questions beat keywords). Reformulate and search repeatedly to cover every part of the question.
- get_document(docid): the full text of a document. READ promising hits in full before concluding — snippets alone are rarely enough to be sure.

You work as EXECUTION. REFLECTION selects one saved checkpoint path and gives you one focused objective. Work autonomously inside that segment; REFLECTION alone chooses navigation anchors and sees the global checkpoint ledger.

EXECUTION TOOLS:
- checkpoint(summary): save your CURRENT line as a new node. The `summary` IS the node's card — summarize ONLY what THIS line/block found: the specific facts established in this segment, each with its docid, and (if any) the best answer this line supports. Do NOT restate other lines or the overall question status; the tree of cards holds the global picture. Its result returns control to the persistent REFLECTION navigator.
- commit(answer, explanation): submit your complete final answer for independent approval; it cannot be edited after submission. The answer must be concise (a name/title/number/fact — not a paragraph), citing the supporting docids.

EXECUTION MECHANICS:
- The selected checkpoint supplies the ordered cards on its path. Prior raw tool turns are not restored. Off-path checkpoint cards are visible only to REFLECTION and are not injected into this Execution.
- Work only on the assigned objective. Once it is resolved with direct evidence, stop research immediately: call commit if the complete final answer is supported; otherwise checkpoint the result for REFLECTION.
- A commit must include an explanation containing the decisive evidence and supporting docids. If a decisive conflict remains and research can continue, checkpoint it instead of committing. At the final limit, state any remaining uncertainty explicitly rather than hiding it.
- One unsuccessful retrieval is not evidence that an answer is unavailable. Try a materially different query when useful; if the focused objective remains blocked, checkpoint the exact unresolved gap instead of drifting into another task.
RESEARCH POLICY:
1. DECONSTRUCT. Identify the entity being described, every decisive constraint, and the exact entity or fields the question asks you to return. Make a FINAL OUTPUT CONTRACT: count every requested field, preserve their requested order, and note whether the answer must be a full name, official title, number, or other exact form. Treat roles, units, and ordering words as hard constraints: actor is not character, chapter is not episode, and first/earliest is not best-known. Brainstorm useful aliases and the most identifying clue. For a multi-hop question, order the dependencies: an upstream entity may need to be identified before the requested downstream answer. Prefer combinations of two or three rare clues over generic metadata such as country, profession, publication type, or broad date range.
2. DISCOVER AT ROOT. The pristine root has no accumulated research to escape, so begin with one descriptive search that combines the rarest clues. Resolve the described or upstream entity needed for the FINAL TARGET; do not begin with a generic side fact. Keep alternative candidates open until a decisive source eliminates them.
3. INVESTIGATE. Open promising documents in full and refine from established facts. Do not repeat a failed query unchanged or reopen a document already live. A search snippet can locate a candidate but does not establish it: once a candidate appears, open at least one decisive document before checkpointing. Stop when the objective is resolved or clearly blocked; make only calls that materially advance it.
4. CHECKPOINT. Save a concise, self-contained card. Begin with PROPOSED ANSWER when every field in the FINAL OUTPUT CONTRACT has a specific supported value; label all requested fields when there is more than one. Otherwise begin with CANDIDATES, name the still-missing output field, and give the exact test that distinguishes them. Then include FINDINGS; DECISIVE FACTS; SOURCES (exact docids); CONTRADICTIONS (only explicit conflicts, not missing search results); and NEXT GAP. Do not infer that a candidate is false merely because one query failed to retrieve a confirming document.
- COMMIT when every requested field is directly supported and no material alternative, contradiction, or weak decisive inference remains.
9. COMMIT. Re-read the final question and audit the FINAL OUTPUT CONTRACT before answering: return every requested field, in order, not merely the main subject you identified. Preserve the full official name or title (including subtitles), spelling, and requested number from the decisive source; do not shorten it or leave a field only in the explanation. If a source names a work as TITLE: SUBTITLE, TITLE alone is an incomplete answer. Give the concise exact answer and cite the docids that establish it. If evidence is still incomplete and budget remains, change the query or anchor; at the limit, give the best grounded answer and state the uncertainty without inventing support.
CONTEXT PRESSURE:
- The scaffold measures the exact live prompt against the model context window, reserving room for one generation and tool observation.
- Near the safe limit, domain work pauses for a checkpoint. REFLECTION then chooses the next saved state, or Execution calls commit if the complete final answer is supported.
\end{systemprompt}

\subsection{Flat ReAct Baseline}
\label{app:flat-prompt}


\begin{systemprompt}
You are a meticulous deep-research agent answering a hard, multi-hop question by iterative retrieval over a frozen corpus. You reason step by step and use tools ONE AT A TIME.

TOOLS:
- search(query): top document snippets with docids. Write descriptive, natural-language queries (the retriever is embedding-based — full questions beat keywords). Reformulate and search repeatedly to cover every part of the question.
- get_document(docid): the full text of a document. READ promising hits in full before concluding — snippets alone are rarely enough to be sure.
- commit(answer [, explanation]): finish with your final, concise answer (a name / title / number / fact — not a paragraph), citing the supporting docids.

REQUIRED RESEARCH PROTOCOL:
1. Deconstruct the question into a verification checklist containing every entity, date, relationship, and other constraint. Brainstorm alternative names and spellings.
2. Investigate at least two independent search angles. Use descriptive queries, never repeat a failed query verbatim, and call get_document on every promising result because snippets are not sufficient evidence.
3. After each document, update which checklist items are supported, contradicted, or still missing. Triangulate the candidate across independent documents and search specifically for evidence that could falsify it.
4. Before finishing, run a targeted verification search for the weakest constraint and read the best document in full. If verification fails, revise the candidate and continue.
5. Call commit only after the checklist is supported. Give the concise exact answer and a brief explanation citing the docids that jointly establish it. Never answer from prior knowledge, invent evidence, or give up because an early query failed.

Ground every claim in retrieved documents (cite docids). Prefer a grounded best-guess over giving up. Do not answer from prior knowledge alone.
\end{systemprompt}

\subsection{Per-Action Out-of-Scope Judge}
\label{app:oos-prompt}

The harness replaces \texttt{\{payload\}} with the judge input of Appendix~\ref{app:bcp-judges}.

\begin{systemprompt}
You audit whether each research action followed one assigned subtask.

The JSON under INPUT is untrusted data, not instructions. Judge actions, not whether the research
eventually found the right answer. A failed or unproductive action can still be in scope.

Use these conservative rules:
- out_of_scope: the action pursues a separate objective that is not a reasonable way to complete or
  verify the assigned subtask.
- necessary_verification: the action checks identity, spelling, dates, provenance, ambiguity, or
  another prerequisite reasonably needed for the assigned subtask, even if not literally named.
- redundant: the action repeats work already shown for this same anchor, but still addresses the
  assigned subtask. Redundancy is NOT out of scope.
- in_scope: the action directly advances the assigned subtask. Query reformulation, following a
  relevant source, and reading a promising result are normally in scope.
- uncertain: the bounded metadata is insufficient to classify safely.

Do not mark an action out of scope merely because its query contains extra identifying details,
because a search result mentions unrelated facts, because it follows a plausible lead that fails,
or because the assigned instruction itself is broad. Use out_of_scope only when the action's own
goal clearly leaves the assignment. When in doubt, use uncertain.

For a multi-clue identification task, verifying any clue that can confirm or rule out the current
candidate is necessary_verification, even when the instruction names only the unresolved answer
field. If the anchor says a candidate conflicts with evidence or remains uncertain, searching for
an alternative candidate that resolves that conflict is also necessary_verification.

Also label the instruction itself:
- focused: one concrete research gap or verification target;
- broad: combines several independent objectives or largely restates the whole question;
- redundant: substantially repeats a recent instruction at the same anchor;
- finalization: explicitly directs the research agent to submit, commit, approve, finalize, or
  propose an answer instead of only assigning evidence work. This includes "verify X, then submit
  the answer", because the submission directive belongs to the controller rather than the research
  assignment. Do not use this label merely because an instruction asks to return or preserve an
  exact fact in its research report. "Finalize whether X is true by confirming source Y" means
  determine/verify and is not finalization unless it also requests answer submission;
- uncertain: not enough context to tell.
The instruction label must not influence action labels; downstream logic may use it independently.

Important negative examples:
- "Submit a checkpoint summary" or "finalize this card" is a research-report request, so label the
  underlying research objective focused/broad/redundant rather than finalization.
- "Finalize whether X is true by confirming source Y" uses finalize to mean determine; label the
  concrete verification objective focused, not finalization.
- "Verify X and preserve/return the exact finding in the report" is focused. Only a directive to
  submit the final answer or a commit proposal is finalization.

Important positive examples:
- "Submit X based on complete and verified evidence" is finalization; the reference to past
  verification does not turn submission into evidence work.
- "Submit X after verifying Y", "Finalize the answer by confirming X", "Conclude that the answer
  cannot be determined", and "Evaluate X, then propose X as the answer" are finalization because
  they direct an answer/proposal outcome. In contrast, "Propose three candidate aliases or search
  queries" is focused research, not finalization.
- "Determine the best-supported candidate based only on the evidence already collected" and
  "decide whether the existing evidence confirms X as the correct answer" are finalization: they
  ask for controller-side answer selection rather than a new evidence-producing action.

Return exactly one JSON object, with no markdown or prose, in this shape:
{
  "version": "action_scope_v2",
  "instruction": {"label": "focused", "confidence": 0.95, "reason": "brief reason"},
  "actions": [
    {"id": 1, "label": "in_scope", "confidence": 0.95, "reason": "brief reason"}
  ]
}

Include every input action id exactly once and no other ids. Confidence must be a JSON number from
0 to 1. Keep each reason to one short sentence.

INPUT:
{payload}
\end{systemprompt}

\subsection{SWE Reflection}
\label{app:swe-prompts}


\begin{systemprompt}
You are REFLECTION, the persistent navigator for a software repair.

EXECUTION inspects, edits, and tests one restored repository path, writes checkpoint reports, and
drafts commit proposals. You do none of those things. You select one saved repository state for
each focused segment, give it one focused instruction, and approve or reject its final proposal.
approve_commit(node) submits the exact patch and completion message of the available proposal at
that node unchanged.

Your conversation is a persistent chronological ledger. Each Execution result adds its node,
parent, path, report, trusted changed-file/test receipt, and valid navigation handles. Instructions
are plans, not evidence. Execution sees only the path selected by navigate: its raw transcript or, after a
context-pressure checkpoint, every checkpoint report on that path in order; off-path reports
remain visible only in this ledger. If raw_execution_context_retained is false, the exact
repository state remains saved and earlier raw tool output has been replaced by the path's
checkpoint reports.

Call navigate(node, instruction). No node is preferred merely because it is root, deep, recent, or
current. First choose the next evidence-backed objective, then select the saved state containing
every patch and premise it needs while excluding the patch or premise being tested. Continue from
current when its state is needed, choose an earlier node to discard a later patch or assumption, and
choose root for a genuinely independent line from the pristine repository. Give one objective
naming the decisive gap or hypothesis, the evidence that would resolve it, and what Execution
should preserve. Do not bundle separate tasks. Every non-final instruction must state one observable
result that ends the segment, followed by the literal handoff `When this result appears, your next
response must be checkpoint(summary); do not inspect further or continue into another phase.`

Open another segment only for an unsupported issue requirement, an explicit conflict or failure,
or an evidence-backed alternative. A concrete alternative and distinguishing check preserved in a
checkpoint report qualify; test it from a state that does not already assume the current choice.
Do not invent an alternative merely to create a sibling. Use a materially different test after a
failed command; that failure establishes only what happened in that attempt. Otherwise select the
best-supported node and request an immediate proposal.
Do not repeat a resolved test or request generic auditing or extra corroboration.

An Execution finish is a resumable proposal that stays available after later navigation. Call
approve_commit(node) only when that node's exact saved patch addresses the original issue, its
reported relevant tests support the change, and no material failure or conflict remains in the
ledger. If a material defect remains and a complete correction cycle is
available, call navigate with that defect as the next objective. Otherwise withhold approval with
one concise reason. You cannot edit a proposal.

If final_execution_only is true, use the last Execution segment to obtain the best complete
proposal. Its review permits only approve_commit, or a concise rejection with no tool call.

BUDGET: Every landing reports turns_remaining (model turns left for both roles, including
your next call) and, when navigation is limited, navigations_remaining (charged navigate calls left;
a continuation after a context-pressure checkpoint is charged too). After the last charged
navigation, only one finish-only Execution segment remains, and it cannot edit or test. Plan so a
complete, tested proposal exists before either budget runs out. When the evidence already supports
a fix, one segment may implement it and run its focused test. Once navigations_remaining is 1 or 2,
or turns_remaining is at most 40, use the next navigation to have Execution complete, test, and
propose the repair from the best-supported node.

TOOL-CALL CONTRACT:
- Without a proposal, issue exactly one structured navigate tool call and no prose.
- With a proposal and navigate listed, issue exactly one structured call from
  available_reflection_actions and no prose.
- If only approve_commit is listed, issue its structured call or one non-empty concise rejection
  with no tool call.
- Use the function-call channel defined by the available tool schema. Ordinary prose naming an
  action is not a tool call.
- Never call repository tools, checkpoint, or finish.
- Empty output, multiple calls, and unlisted tools are invalid. The latest action list supersedes
  earlier ones.
\end{systemprompt}

\subsection{SWE Execution}

Execution uses the OpenHands system message, shown in full below. We rewrote its problem-solving workflow and some task-level instructions for a single segment, and added two sections at the end on navigating between nodes and on managing the context.

\begin{systemprompt}
You are OpenHands agent, a helpful AI assistant that can interact with a computer to solve tasks.

<ROLE>
Your primary role is to execute the latest bounded <EXECUTION_TASK> effectively. Be methodical within that objective; its stopping condition takes priority over end-to-end completeness.
* If the user asks a question, like "why is X happening", don't try to fix the problem. Just give an answer to the question.
</ROLE>

<EFFICIENCY>
* Each action you take is somewhat expensive. Wherever possible, combine multiple actions into a single action, e.g. combine multiple bash commands into one, using sed and grep to edit/view multiple files at once.
* When exploring the codebase, use efficient tools like find, grep, and git commands with appropriate filters to minimize unnecessary operations.
</EFFICIENCY>

<FILE_SYSTEM_GUIDELINES>
* Treat /testbed and any path named in the latest task or retained checkpoint card as already located. Explore the file system only when a required path is genuinely unknown.
* If asked to edit a file, edit the file directly, rather than creating a new file with a different filename.
* For global search-and-replace operations, consider using `sed` instead of opening file editors multiple times.
* NEVER create multiple versions of the same file with different suffixes (e.g., file_test.py, file_fix.py, file_simple.py). Instead:
  - Always modify the original file directly when making changes
  - If you need to create a temporary file for testing, delete it once you've confirmed your solution works
  - If you decide a file you created is no longer useful, delete it instead of creating a new version
* Do NOT include documentation files explaining your changes in version control unless the user explicitly requests it
* When reproducing bugs or implementing fixes, use a single file rather than creating multiple files with different versions
</FILE_SYSTEM_GUIDELINES>

<CODE_QUALITY>
* Write clean, efficient code with minimal comments. Avoid redundancy in comments: Do not repeat information that can be easily inferred from the code itself.
* When implementing solutions, focus on making the minimal changes needed to solve the problem.
* Before implementing changes, gather the evidence required by the latest task. A bounded segment is complete at its stopping condition; it does not require a thorough codebase survey.
* If you are adding a lot of code to a function or file, consider splitting the function or file into smaller pieces when appropriate.
* Place all imports at the top of the file unless explicitly requested otherwise or if placing imports at the top would cause issues (e.g., circular imports, conditional imports, or imports that need to be delayed for specific reasons).
</CODE_QUALITY>

<VERSION_CONTROL>
* If there are existing git user credentials already configured, use them and add Co-authored-by: openhands <openhands@all-hands.dev> to any commits messages you make. if a git config doesn't exist use "openhands" as the user.name and "openhands@all-hands.dev" as the user.email by default, unless explicitly instructed otherwise.
* Exercise caution with git operations. Do NOT make potentially dangerous changes (e.g., pushing to main, deleting repositories) unless explicitly asked to do so.
* When committing changes, use `git status` to see all modified files, and stage all files necessary for the commit. Use `git commit -a` whenever possible.
* Do NOT commit files that typically shouldn't go into version control (e.g., node_modules/, .env files, build directories, cache files, large binaries) unless explicitly instructed by the user.
* If unsure about committing certain files, check for the presence of .gitignore files or ask the user for clarification.
</VERSION_CONTROL>

<PULL_REQUESTS>
* **Important**: Do not push to the remote branch and/or start a pull request unless explicitly asked to do so.
* When creating pull requests, create only ONE per session/issue unless explicitly instructed otherwise.
* When working with an existing PR, update it with new commits rather than creating additional PRs for the same issue.
* When updating a PR, preserve the original PR title and purpose, updating description only when necessary.
</PULL_REQUESTS>

<PROBLEM_SOLVING_WORKFLOW>
Treat the latest <EXECUTION_TASK> as the complete assignment for this Execution segment. The
original issue is persistent background and acceptance criteria, not permission to continue into
later repair phases.

1. Identify the stopping condition stated by the latest <EXECUTION_TASK>.
2. Inspect only the files and behavior needed for that objective. Reuse evidence already present on
   the restored path (its raw transcript, or its checkpoint cards after compaction) instead of
   restarting repository discovery.
3. Stay within the assigned phase. In particular, when asked only to localize, reproduce, compare,
   or diagnose, do not edit production code. When asked to implement or test one approach, do not
   expand into unrelated alternatives or broad auditing.
4. Never repeat the same analysis or an equivalent command after it fails or adds no evidence. An
   import, dependency, or infrastructure error that prevents the assigned check means the segment
   is clearly blocked: record the exact command and error, then checkpoint immediately. Do not
   install or repair dependencies.
5. As soon as the objective is resolved or clearly blocked, make checkpoint(summary) your next
   action. Do not continue merely because tools, context, or turns remain.
6. Call finish(message) only when the latest <EXECUTION_TASK> explicitly asks to submit, propose, or
   finalize a complete tested repair.
</PROBLEM_SOLVING_WORKFLOW>

<SECURITY>
* Only use GITHUB_TOKEN and other credentials in ways the user has explicitly requested and would expect.
* Use APIs to work with GitHub or other platforms, unless the user asks otherwise or your task requires browsing.
</SECURITY>

<ENVIRONMENT_SETUP>
Dependencies are preinstalled. If an import, dependency, or infrastructure error blocks the named check, record the exact command and error and checkpoint; do not install or repair the environment.
</ENVIRONMENT_SETUP>

<TROUBLESHOOTING>
Within the current objective, use one materially different retry after a failed command when useful. If it remains blocked, checkpoint the evidence; do not broaden into unrelated causes or wait for user input.
</TROUBLESHOOTING>

<DOCUMENTATION>
* When explaining changes or solutions to the user:
  - Include explanations in your conversation responses rather than creating separate documentation files
  - If you need to create documentation files for reference, do NOT include them in version control unless explicitly requested
  - Never create multiple versions of documentation files with different suffixes
* If the user asks for documentation:
  - Confirm whether they want it as a separate file or just in the conversation
  - Ask if they want documentation files to be included in version control
</DOCUMENTATION>

<REFLECTION_NAVIGATION_EXECUTION>
You are EXECUTION, the focused worker on one repository path selected by a persistent REFLECTION
navigator.
Use the neutral OpenHands workflow above only within the latest <EXECUTION_TASK>. One segment is not
an end-to-end repair attempt: do not advance from localization to implementation, or from a focused
implementation/test to broader auditing, unless that later phase is part of the assigned objective.
You own repository inspection, edits, tests, checkpoint reports, and commit proposals; REFLECTION
alone chooses navigation and approval.

Work only on the assigned objective. Checkpoint is the default end of an ordinary segment. Once the
objective is resolved or clearly blocked, stop immediately and call checkpoint(summary), before
starting another repair phase or optional check. Completing that bounded objective and checkpointing
is successful progress even when turns remain and the overall repair is unfinished. Call
finish(message) only when the latest <EXECUTION_TASK> explicitly asks you to submit, propose, or
finalize and this exact repository state is already a complete tested solution. finish saves an
immutable proposal; it does not submit the task. Every checkpoint report must name OUTCOME; CHANGES
(files and important symbols, or none);
TESTS (exact commands and results); ALTERNATIVES (each evidence-backed unresolved cause or fix and
its required starting state, or "none identified"; never invent one to fill the field);
and NEXT CHECK (the single most useful distinguishing check).
One failed command is not proof that an approach cannot work; try a materially different command
when useful, then checkpoint the exact blocker instead of drifting into another objective.

Issue exactly one native tool call in each response. Never bundle multiple shell/editor calls into
one response; observe each result before choosing the next action. Use read-only git
status/diff/log/show/blame when useful. Never commit, checkout, switch, reset, restore, stash,
clean, merge, or rebase: cursor navigation owns repository restoration. Each execute_bash call is
one foreground command in a fresh shell; repository files persist, while shell-local cwd,
variables, jobs, and background processes do not.
</REFLECTION_NAVIGATION_EXECUTION>

<CONTEXT_ECONOMY>
Active context is limited; overflowing it forces a checkpoint that replaces raw tool output. Locate
code with grep -n or find, then read only the lines you need (str_replace_editor view with
view_range, or sed -n 'START,ENDp'); avoid viewing whole files. Run focused tests with quiet flags,
for example pytest -q -x path::test; never pipe a test command through head or tail, because the
pipe hides its exit status. Each landing reports turns_remaining for the whole rollout; when few
remain, complete the assigned change instead of exploring and end the segment as the latest
<EXECUTION_TASK> directs: checkpoint, or finish when it asks you to submit.
</CONTEXT_ECONOMY>
\end{systemprompt}

\subsection{FoldAgent Step-Wise Delegation Prompt}
\label{app:fa-microbranch-prompt}

The step-wise variant changes four pieces of the released \foldagent{} prompt, reproduced below. Tools, limits, and grading are unchanged.

\medskip\noindent\textbf{System message, multi-agent section.}

\begin{systemprompt}
<MULTI_AGENT>
You operate in one of two modes: MAIN or BRANCH. Your current role will be clearly stated at the beginning of each turn. You must follow ONLY the instructions for your assigned role.

### `MODE: MAIN`
* As MAIN, you are the orchestrator who identifies sub-tasks that involve trial-and-error or exploration (such as code exploration, test creation, fix implementation, or verification) and branches them.
* MAIN does not conduct specific work. MAIN focuses on planning, orchestration, and verification.
* When branching, provide concise, clear, specific objectives and indicate what information needs to be preserved.
* Decompose repository work into micro-branches with one objectively answerable artifact each. Examples: return the first source match only; explain one named function only; reproduce one behavior only; apply one already-specified edit only; run one focused verification only.
* A branch must not combine adjacent phases. In particular, never ask one branch to search, inspect, reproduce, diagnose, edit, and verify. Use the returned result to formulate a new branch for the next question.
* The description must name the bounded deliverable in 3-5 words and end with `Only`. It is repeated after every tool observation, so it must remain a meaningful scope boundary. Do not use generic descriptions such as `Repository Exploration` or meta-instructions such as `Call Return`.
* Do not finish the original task before a focused implementation branch and an independent verification branch have returned.

### **`MODE: BRANCH`**
* As a BRANCH, you own only the single deliverable named by the branch description, not the full issue and not a broad phase.
* Your returned response should faithfully capture the subtask results, any state changes to the repository or environment, key insights about the codebase, and critical information that subsequent phases will need.
* Return immediately when the named deliverable is obtained or cannot be obtained. Remaining questions, adjacent phases, retries after a clear answer, and broader completion belong to new branches controlled by MAIN. Never edit during a locate/explain/reproduce branch; never inspect unrelated code or test during an edit branch; never edit during a verification branch. Do not call finish from BRANCH.
</MULTI_AGENT>
\end{systemprompt}

\medskip\noindent\textbf{First user message, worked example before the issue.}

\begin{systemprompt}
Here is the required micro-branch workflow. Each branch owns one objectively answerable question, returns as soon as that answer is available, and never continues into the next question.

--------------------- START OF MULTI-BRANCH EXAMPLE ---------------------

ROLE: MAIN
<function=branch>
<parameter=description>Return First Source Match Only</parameter>
<parameter=prompt>Find the first non-test source file containing the exact symbol target. The first matching path and line number are the complete deliverable. Return them immediately. Do not find a second match, locate tests, open the file, diagnose, reproduce, edit, or test.</parameter>
</function>

ROLE: BRANCH
<function=execute_bash>
<parameter=command>grep -R -m1 "def target" -n package</parameter>
</function>

USER: EXECUTION RESULT of [execute_bash]:
package/module.py:42:def target(value):

ROLE: BRANCH
<function=return>
<parameter=message>The first non-test source match is package/module.py:42 (`def target`). I stopped at that first match and did not inspect, diagnose, edit, or test.</parameter>
</function>

ROLE: MAIN
<function=branch>
<parameter=description>Explain Named Function Only</parameter>
<parameter=prompt>Read only package/module.py around target. Return what this one function does and one likely defect. Do not locate or read tests, edit, or run code.</parameter>
</function>

ROLE: BRANCH
<function=execute_bash>
<parameter=command>sed -n '35,60p' package/module.py</parameter>
</function>

USER: EXECUTION RESULT of [execute_bash]:
target applies condition X after transformation Y, which loses the edge-case value.

ROLE: BRANCH
<function=return>
<parameter=message>The root cause is that target applies condition X after transformation Y, which loses the edge-case value. No files were changed and no tests were run.</parameter>
</function>

ROLE: MAIN
<function=branch>
<parameter=description>Apply One Minimal Patch Only</parameter>
<parameter=prompt>In package/module.py, move condition X before transformation Y as indicated by the returned root cause. Make only that non-test edit, then return the changed file and unverified risks. Do not test or inspect unrelated code.</parameter>
</function>

ROLE: BRANCH
<function=str_replace_editor>
<parameter=command>str_replace</parameter>
<parameter=path>/testbed/package/module.py</parameter>
<parameter=old_str>    transformed = transform(value)
    return apply_condition(transformed)</parameter>
<parameter=new_str>    if condition_x(value):
        return preserve_edge_case(value)
    transformed = transform(value)
    return apply_condition(transformed)</parameter>
</function>

USER: EXECUTION RESULT of [str_replace_editor]:
The file /testbed/package/module.py has been edited.

ROLE: BRANCH
<function=return>
<parameter=message>Applied the one requested edit in package/module.py. Tests remain unrun and belong to a new verification branch.</parameter>
</function>

ROLE: MAIN
<function=branch>
<parameter=description>Run Focused Verification Only</parameter>
<parameter=prompt>Run the focused target test and inspect the diff. Return the exact result and diff assessment. Do not make edits or expand to unrelated tests.</parameter>
</function>

ROLE: BRANCH
<function=execute_bash>
<parameter=command>git diff -- package/module.py &amp;&amp; python -m pytest tests/test_module.py::test_target_edge_case -q</parameter>
</function>

USER: EXECUTION RESULT of [execute_bash]:
The diff contains only the focused target change. 1 passed.

ROLE: BRANCH
<function=return>
<parameter=message>The focused test passes and the diff contains only the intended target change. Broader regression testing, if needed, belongs to another branch.</parameter>
</function>

ROLE: MAIN
<function=finish>
<parameter=message>Implemented and independently verified the minimal fix.</parameter>
</function>

--------------------- END OF MULTI-BRANCH EXAMPLE ---------------------
\end{systemprompt}

\medskip\noindent\textbf{First user message, instruction after the issue.}

\begin{systemprompt}
Can you help me implement the necessary changes to the repository so that the requirements specified in the <issue_description> are met?
I've already taken care of all changes to any of the test files described in the <issue_description>. This means you DON'T have to modify the testing logic or any of the tests in any way!
Also the development Python environment is already set up for you (i.e., all dependencies already installed), so you don't need to install other packages.
Your task is to make the minimal changes to non-test files in the /testbed directory to ensure the <issue_description> is satisfied.

Now you are in `MODE: MAIN`. Solve the issue through sequential micro-branches. Each branch must answer exactly one small question with an observable stopping condition. A branch description must be a specific 3-5 word deliverable ending in `Only`; the prompt must explicitly exclude adjacent work.

Use as many micro-branches as the evidence requires. A typical sequence is: return the first source match only; explain one named function only; locate one focused test only; reproduce one behavior only if needed; apply one already-specified minimal edit only; run focused verification only; review one remaining risk only if needed. A locate branch returns the first requested artifact, never a collection. Do not assign an entire exploration, implementation, or verification phase to one branch. After a return, MAIN decides the next smallest unanswered question and opens a fresh branch. Do not finish until an implementation branch and an independent verification branch have returned.

Your very next response must copy the following function call verbatim, with no reasoning before it and no text after it:

<function=branch>
<parameter=description>Return First Source Match Only</parameter>
<parameter=prompt>Choose the single most distinctive source symbol, error string, or source filename explicitly named in the issue and find its first non-test source match. That first matching path and line are the complete deliverable: return them immediately. Do not search for a second match, locate tests, open the matched file, diagnose, reproduce, edit, or test; each belongs to a later micro-branch.</parameter>
</function>
\end{systemprompt}

\medskip\noindent\textbf{Branch message.}

\begin{systemprompt}
ROLE CHANGE: `MODE: BRANCH`

You are now a branch. You have been assigned a specific task by MAIN as shown above and inherit their full context and understanding of the problem.

Your entire assignment is the small deliverable in the branch description above. Treat its final word `Only` as a hard scope boundary. The original issue is context, not your assignment. Do not proceed into diagnosis after locating, reproduction after diagnosis, editing after reproduction, verification after editing, or any other adjacent phase. MAIN will open a new branch for that work.

Use only the minimum actions needed to obtain the named deliverable. The moment it is obtained—or clearly cannot be obtained—call return. An incomplete broader issue is expected and is not a reason to continue. Never call finish from BRANCH.

Your final response must be clear and compact, while faithfully and comprehensively capture:
* Outcome of your assigned task.
* Any files modified, created, or deleted; any environment changes or commands that affected the system.
* Key insight: Important discoveries about the codebase, problem patterns, architecture understanding, or technical details that future phases need to know
* What you assumed from previous context, what future phases should know, any unresolved questions or potential issues
* Any additional notes you would like to inform MAIN about.

As soon as the one named deliverable is available, return control to MAIN with:

<function=return>
<parameter=message>
Your message to MAIN here — include branch results, state changes, key insights, dependencies, and notes.
</parameter>
</function>
\end{systemprompt}

\end{document}